\documentclass[11pt]{article}

\usepackage[final]{acl}
\usepackage{times}
\usepackage{latexsym}
\usepackage[T1]{fontenc}
\usepackage[utf8]{inputenc}
\usepackage{microtype}
\usepackage{inconsolata}
\usepackage{tabularx}
\usepackage{placeins}

\usepackage{graphicx}
\usepackage{subcaption}

\usepackage{amsmath,amssymb,amsthm,mathtools}
\usepackage{booktabs}
\usepackage{multirow}
\usepackage{siunitx}
\usepackage{array}
\usepackage{makecell}
\usepackage{adjustbox}

\usepackage{algorithm}
\usepackage{algpseudocode}
\usepackage{listings}
\usepackage{float}
\usepackage{tcolorbox}
\tcbuselibrary{listings,skins,breakable}

\definecolor{promptbg}{RGB}{252,252,255}
\definecolor{promptframe}{RGB}{100,149,237}
\definecolor{promptbar}{RGB}{65,105,225}
\definecolor{prompttitle}{RGB}{25,25,112}

\newtcolorbox{promptbox}[1][]{
  enhanced,
  breakable,
  colback=promptbg,
  colframe=promptframe,
  coltitle=prompttitle,
  fonttitle=\bfseries\sffamily\small,
  attach boxed title to top left={yshift=-2mm, xshift=4mm},
  boxed title style={
    colback=white,
    colframe=promptframe,
    boxrule=0.5pt,
    arc=2pt,
  },
  borderline west={3pt}{0pt}{promptbar},
  left=6pt,
  right=6pt,
  top=8pt,
  bottom=6pt,
  boxrule=0.5pt,
  arc=4pt,
  shadow={1mm}{-1mm}{0mm}{black!15},
  #1
}

\usepackage[table]{xcolor}
\usepackage{colortbl}
\usepackage{enumitem}
\usepackage{wrapfig}
\usepackage{pifont}
\usepackage{url}
\definecolor{stdgray}{gray}{0.45}

\newcommand{\cmark}{\ding{51}}
\newcommand{\xmark}{\ding{55}}

\definecolor{lightpurple}{RGB}{245,240,255}
\definecolor{lightblue}{RGB}{235,245,255}
\definecolor{darkgreen}{RGB}{0,100,0}
\newcolumntype{L}[1]{>{\raggedright\arraybackslash}p{#1}}
\newcolumntype{C}[1]{>{\centering\arraybackslash}p{#1}}

\DeclareMathOperator*{\argmax}{arg\,max}
\newcommand{\E}{\mathbb{E}}

\newcommand{\errorguided}{\textsc{Error-Guided}}

\title{Adaptive Prompt Structure Factorization:\\
A Framework for Self-Discovering and Optimizing\\
Compositional Prompt Programs}

\author{
Haoyue Liu\textsuperscript{$\spadesuit$}\quad
Zhichao Wang\textsuperscript{$\spadesuit$}\quad
Yongxin Guo\textsuperscript{$\heartsuit$}\quad
Haoran Shou\textsuperscript{$\spadesuit$}\quad
Xiaoying Tang\textsuperscript{$\spadesuit$*}\\[6pt]
\textsuperscript{$\spadesuit$}The Chinese University of Hong Kong, Shenzhen\quad
\textsuperscript{$\heartsuit$}Taobao and Tmall Group\\[3pt]
\texttt{\{haoyueliu, zhichaowang, haoranshou\}@link.cuhk.edu.cn}\\
\texttt{guoyongxin.gyx@taobao.com}\quad
\texttt{tangxiaoying@cuhk.edu.cn}
}

\begin{document}
\maketitle

\renewcommand{\thefootnote}{\fnsymbol{footnote}}
\footnotetext[1]{Corresponding author.}
\footnotetext[2]{Code available at \url{https://github.com/luckyboyLiu/aPSF}}
\renewcommand{\thefootnote}{\arabic{footnote}}

\begin{abstract}
Automated prompt optimization is crucial for eliciting reliable reasoning from large language models (LLMs), yet most API-only prompt optimizers iteratively edit monolithic prompts, coupling components and obscuring credit assignment, limiting controllability, and wasting tokens. We propose \textbf{Adaptive Prompt Structure Factorization (aPSF)}, an \textbf{API-only} framework (prompt-in/text-out; no access to model internals) that uses an \emph{Architect} model to discover task-specific prompt structures as semantic factors. aPSF then performs interventional, single-factor updates: \emph{interventional factor-level scoring} estimates each factor's marginal contribution via validation-performance changes, and \emph{error-guided factor selection} routes updates to the current dominant failure source for more sample-efficient optimization. Across multiple advanced reasoning benchmarks, aPSF outperforms strong baselines including principle-aware optimizers, improving accuracy by up to \textbf{+2.16} percentage points on average, and reduces optimization cost by \textbf{45--87\%} tokens on MultiArith while reaching peak validation in \textbf{1} step.

\end{abstract}

\section{Introduction}
\label{sec:intro}

\begin{figure}[t]
\centering
\IfFileExists{Fig/MOTIVATION-new.png}{%
  \includegraphics[width=\linewidth]{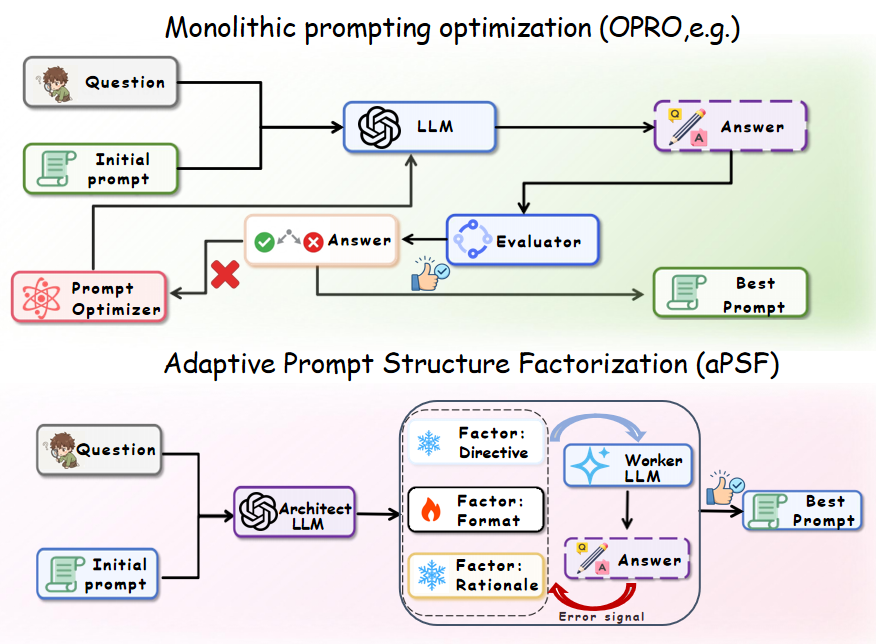}%
}{%
  \includegraphics[width=\linewidth]{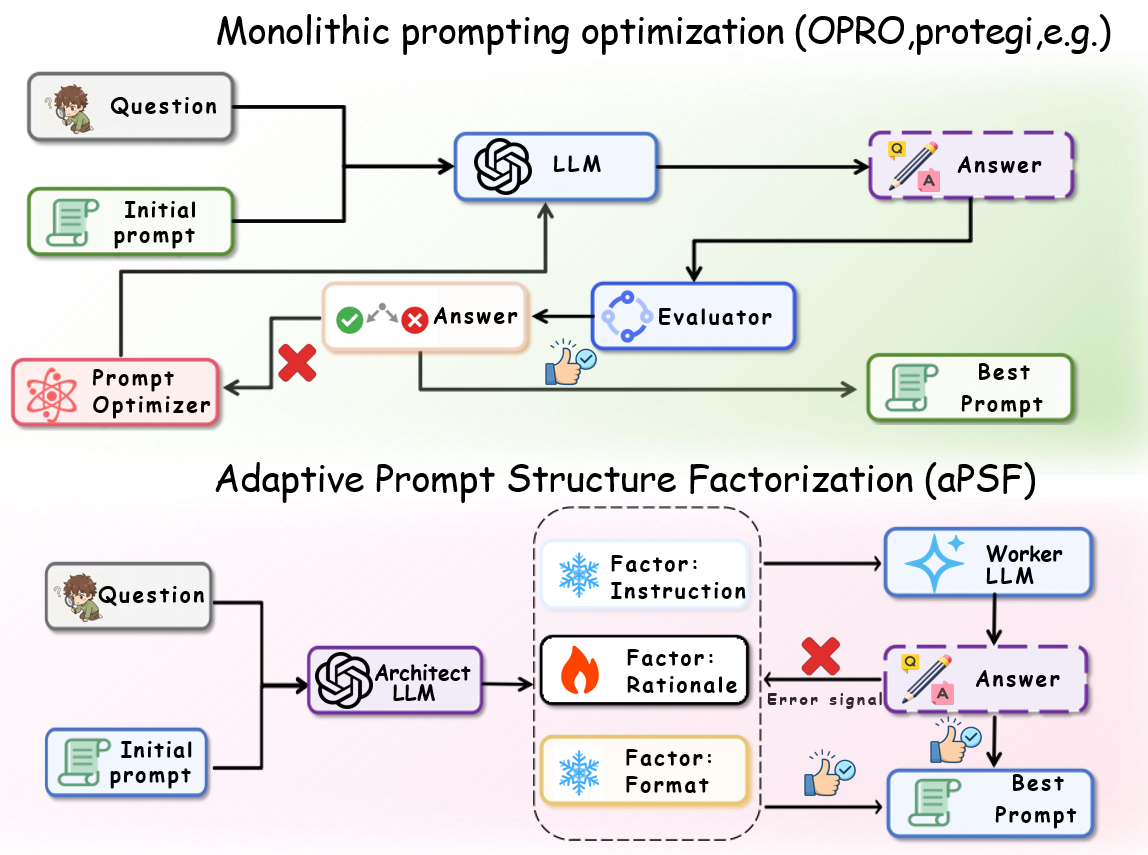}%
}
\caption{
\textbf{Monolithic prompt optimization vs.\ aPSF.}
Top: monolithic API-only prompt optimizers iteratively edit a single prompt.
Bottom: aPSF decomposes the prompt into semantic factors and updates selected factors while freezing the rest.
}

\label{fig:motivation}
\end{figure}

Large language models (LLMs) exhibit strong capabilities in reasoning, coding, and complex generation~\citep{achiam2023gpt,wei2022chain,kojima2022large,wang2022self,brown2020language}, yet their performance is highly sensitive to prompt design~\citep{zhao2021calibrate,liu2023pre}. Under API-only (query-only) access, prompt optimizers typically improve performance by iteratively editing a prompt and selecting candidates by validation score~\citep{zhou2022large,yang2023large,pryzant2023automatic,fernando2023promptbreeder,guo2023connecting,deng2022rlprompt,tang2025unleashing}. Early approaches such as APE~\citep{zhou2022large} and OPRO~\citep{yang2023large} treat the prompt as a holistic, indivisible string, proposing and selecting entirely new prompts at each iteration. Figure~\ref{fig:motivation} contrasts monolithic prompt editing with aPSF's factor-wise optimization. However, prompt-level edits to a monolithic prompt have three limitations. First, coupled edits obscure which prompt component caused an improvement, complicating credit assignment. Second, monolithic prompts offer weak controllability, preventing selective freezing of constraint-critical parts such as output schemas. Third, because a single mutation often perturbs multiple functional roles at once, many trials are confounded and rejected for unrelated reasons, wasting query budget and slowing convergence.

In practice, effective prompts often behave like \emph{compositional programs} with semantic factors (e.g., task interpretation, reasoning procedure, output format, verification)~\citep{white2023prompt}. Edit-based optimizers such as ProTeGi~\citep{pryzant2023automatic} and GrIPS~\citep{prasad2023grips} partially address this by applying localized mutations. However, without an explicit, persistent factorization, their optimization and selection still operate at the level of the full prompt, leaving credit assignment implicit and limiting selective freezing. These limitations motivate our central question: \emph{Can we automatically induce a task-specific prompt factorization and optimize factors independently, enabling explicit credit assignment and adaptive allocation of the query budget?}

To bridge this gap, we introduce \emph{Adaptive Prompt Structure Factorization (aPSF)}, a two-phase framework for API-only prompt optimization that incorporates: (1) \textbf{adaptive structure discovery}, where an Architect LLM induces a task-specific factor schema and initializes factor contents (optionally conditioned on a user-provided initial prompt); and (2) \textbf{interventional factor optimization}, where we edit one factor at a time and use error-guided factor selection to route updates to the current bottleneck factor.

\paragraph{Contributions.}
Building on this framework, our contributions are fourfold.
\textbf{(1)} We first decompose prompts into task-specific semantic factors, enabling precise, surgical single-factor edits while freezing the remaining factors. Under strictly API-only access, this factorization enables factor-level meta-attribution, effectively \textbf{opening the black box of prompt engineering}.
\textbf{(2)} We propose \emph{interventional factor-level scoring} to quantify the marginal contribution of each component, transforming prompt optimization into a \textbf{traceable diagnostic process} that directly maps refinement decisions to observed validation failures.
\textbf{(3)} We develop an \emph{error-guided factor selection} strategy that adaptively routes the black-box query budget to identified bottleneck factors, thereby \textbf{maximizing sample efficiency and optimization stability}.
\textbf{(4)} Across six reasoning benchmarks, aPSF consistently outperforms strong baselines and yields more stable and interpretable optimization trajectories.

\section{Related Work}
\label{sec:related_work}

\paragraph{API-only prompt optimization.}
Gradient-based methods such as \textbf{AutoPrompt}~\citep{shin2020autoprompt}, \textbf{Prompt Tuning}~\citep{lester2021power}, and \textbf{Prefix-Tuning}~\citep{li2021prefix} require white-box access and thus are unsuitable for API-based models.
API-only optimizers therefore rely on prompt-level search and editing, e.g., APE~\citep{zhou2022large}, OPRO~\citep{yang2023large}, ProTeGi~\citep{pryzant2023automatic}, and GrIPS~\citep{prasad2023grips}.
While candidate generation can be holistic or localized, these methods typically operate on a monolithic prompt and select candidates at the whole-prompt level, leaving component-level credit assignment implicit and limiting controllability.
Recent principle-aware methods such as CriSPO~\citep{he2025crispo} and ZERA~\citep{yi2025zera} improve upon monolithic editing by generating structured critiques or evolving principles to guide prompt refinement. However, these principles serve as \emph{evaluation criteria} for feedback generation rather than an explicit \emph{edit space}: the optimization still proposes and selects whole-prompt candidates, limiting factor-level attribution and selective freezing.
To address these limitations without changing the API-only setting, aPSF introduces an explicit factorized representation and optimizes via interventional, single-factor substitutions, making improvements attributable and controllable, even under tight query budgets.

\paragraph{Programmatic prompting and modular pipelines.}
Frameworks such as DSPy~\citep{khattab2024dspy} and SAMMO~\citep{schnabel2024prompts} optimize prompts and demonstrations within a user-specified program scaffold, while modularization/faceting work studies decomposition under predefined taxonomies or rules~\citep{yao2023tree,juneja2025task}.
These approaches improve prompts within a designer-provided scaffold, but optimization and selection remain end-to-end at the program level, yielding limited component-wise attribution under query-only feedback.
In contrast, aPSF automatically induces the scaffold as semantic factors and optimizes via single-factor interventions with explicit attribution, enabling selective freezing (e.g., output format).

\paragraph{Related optimization techniques.}
Bandit-style allocation~\citep{auer2002finite} and self-refinement methods~\citep{madaan2023selfrefineiterativerefinementselffeedback,shinn2023reflexion,chen2024prompt} motivate adaptive routing and iterative improvement.
However, most prior approaches treat the prompt/output as the unit of action, making it difficult to localize which component should be revised under query-only signals.
aPSF instead routes updates over \emph{prompt factors} via error-guided selection and applies interventional single-factor edits.

\section{Adaptive Prompt Structure Factorization}
\label{sec:method}

\begin{figure*}[t]
    \centering
    \IfFileExists{Fig/new-overview-new.png}{%
      \includegraphics[width=\textwidth]{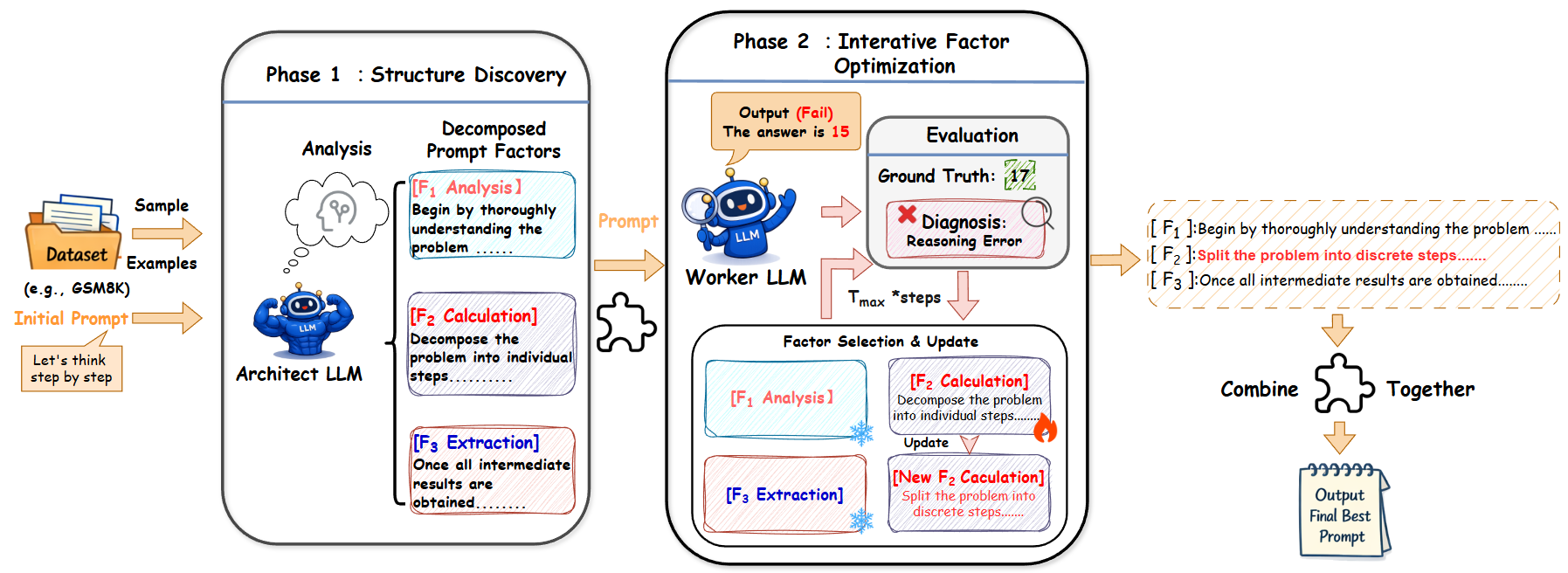}%
    }{%
      \includegraphics[width=\textwidth]{Fig/new-overview-new.png}%
    }
    \caption{
\textbf{Overview of aPSF.}
Given a dataset of sampled examples and an optional initial prompt, aPSF proceeds in two phases.
\textbf{(1) Structure Discovery:} the Architect LLM analyzes the task and decomposes the prompt into $K$ semantic factors $\{F_k\}_{k=1}^K$ (task-specific factorization, rather than a fixed scaffold).
\textbf{(2) Iterative Factor Optimization:} the Worker LLM executes the composed prompt; an evaluator returns scores and representative failure cases, which are summarized into an error diagnosis (e.g., \emph{reasoning error}: \texttt{15} vs\ \texttt{17}).
Conditioned on the diagnosis, the architect model selects a bottleneck factor for a \emph{single-factor} update while freezing the remaining factors, yielding attributable, interventional prompt edits and returning the best-performing prompt within $T_{\max}$ steps.
}

    \label{fig:apsf_overview}
\end{figure*}

We now detail aPSF's factorized prompt representation and optimization procedure (Figure~\ref{fig:apsf_overview}); pseudocode for Phase~2 is provided in Algorithm~\ref{alg:phase2} (Appendix~\ref{app:algorithms}).

\subsection{Problem Setup and Notation}
\label{sec:method:prelim}

Let $\mathcal{D}_{\mathrm{val}}$ and $\mathcal{D}_{\mathrm{test}}$ denote the validation and test sets, where each example is a pair $(q,y)$ of query $q$ and ground-truth answer $y$.
We denote by $p$ a prompt, by $M_W(q;p)$ the Worker LLM's output on $q$ under $p$, and by $s(\cdot,\cdot)$ a task-specific scoring function.
We denote by $T$ the task description and by $\mathcal{X}$ optional exemplars.
\paragraph{Standard objective.}
Most API-only prompt optimizers cast prompt search as maximizing a validation score.
We define the empirical score of prompt $p$ on dataset $\mathcal{D}$ as
\begin{equation}
\label{eq:empirical_score}
\hat{S}(p,\mathcal{D}) \triangleq \frac{1}{|\mathcal{D}|}\sum_{(q,y)\in\mathcal{D}} s(M_W(q;p),y).
\end{equation}
We optimize prompts on $\mathcal{D}_{\mathrm{val}}$ via query-only evaluation and report results on the held-out $\mathcal{D}_{\mathrm{test}}$:
\begin{equation}
\label{eq:objective}
p^\star=\argmax_{p}\; \hat{S}(p,\mathcal{D}_{\mathrm{val}}).
\end{equation}

\paragraph{Traditional monolithic optimization and limitations.}
Existing API-only prompt optimizers typically treat $p$ as a single string and apply global edits~\citep{zhou2022large,yang2023large,pryzant2023automatic,prasad2023grips}.
At step $t$, $\mathrm{Edit}(p^{(t)};T,\mathcal{X})$ proposes candidates $\mathcal{C}^{(t)}$ and updates
$p^{(t{+}1)}=\argmax_{c\in\mathcal{C}^{(t)}} \hat{S}(c,\mathcal{D}_{\mathrm{val}})$.
This design has two limitations: (1) global edits couple semantic components, obscuring credit assignment; (2) coupling prevents selectively freezing components (e.g., output format) and induces a large, unstructured search space.

\paragraph{Addressing the limitations via factorization.}
To address the above limitations, we represent a prompt $p$ as an ordered list of $K$ (task-adaptive) semantic factors.
Let $\mathcal{G}=(\mathcal{F}_1,\ldots,\mathcal{F}_K)$ denote the ordered factor schema, where each $\mathcal{F}_k$ specifies one factor role, and let $B=(f_1,\ldots,f_K)$ be the corresponding factor texts.
We assemble $p$ by concatenation in schema order:
\begin{equation}
\label{eq:assemble}
p = \mathrm{Assemble}(B) \triangleq \mathrm{concat}(f_1, f_2, \ldots, f_K),
\end{equation}
where $\mathrm{concat}(\cdot)$ joins the text blocks with newline delimiters.

\paragraph{Inducing task-specific factors.}
Instead of committing to a fixed taxonomy, we use an Architect LLM $M_A$ to \textbf{induce task-specific semantic factors} from $(T,\mathcal{X})$.
In principle, factorized prompt optimization jointly chooses schema $\mathcal{G}$ and contents $B$:
\begin{equation}
\label{eq:factor_discovery}
\mathcal{G}^*, B^* = \argmax_{\mathcal{G}, B} \hat{S}(\mathrm{Assemble}(B), \mathcal{D}_{\mathrm{val}}).
\end{equation}
This joint search is intractable; we instead obtain an initialization $(\mathcal{G}, B^{(0)})$ via one-pass meta-prompted generation from $M_A$ (Appendix~\ref{app:metaprompts}).

\subsection{Phase 1: Adaptive Structure Discovery}
\label{sec:method:phase1}

In Phase~1, the Architect LLM induces an ordered factor schema $\mathcal{G}$ and initializes the factor contents $B^{(0)}$.
We support two operational modes for structure discovery:

\paragraph{From-scratch mode.}
Given task description $T$ and optional exemplars $\mathcal{X}$, the Architect LLM $M_A$ autonomously proposes a factor schema $\mathcal{G}$ (defining roles and sequence) and initial content $B^{(0)}$:
\begin{align}
\label{eq:struct}
\mathcal{G} &= M_A(p_{\mathrm{struct}}^{\mathrm{meta}}; T, \mathcal{X}), \\
\label{eq:init}
B^{(0)} &= M_A(p_{\mathrm{factorinit}}^{\mathrm{meta}}; \mathcal{G}, T, \mathcal{X}).
\end{align}
We denote by $p_{\mathrm{struct}}^{\mathrm{meta}}$ and $p_{\mathrm{factorinit}}^{\mathrm{meta}}$ the meta-prompts for schema induction and factor initialization (Appendix~\ref{app:metaprompts}).

\paragraph{Initial-prompt mode.}
Let $p_{\mathrm{init}}$ denote an initial prompt provided by the user (e.g., ``Let's think step by step'').
aPSF analyzes its core reasoning approach and generates a task-specific extension. The Architect receives:
\begin{equation}
\label{eq:init_prompt}
\mathcal{G}, B^{(0)} = M_A(p_{\mathrm{analyze}}^{\mathrm{meta}}; p_{\mathrm{init}}, T, \mathcal{X}),
\end{equation}
where $p_{\mathrm{analyze}}^{\mathrm{meta}}$ instructs the architect to: (1)~understand the core reasoning approach of $p_{\mathrm{init}}$, (2)~generate a complete task-specific instruction embodying and extending this approach, and (3)~decompose it into independently optimizable factors.

The meta-prompt additionally encourages preserving the initial prompt's reasoning style while producing a fluent, compact factorization (Appendix~\ref{app:metaprompts}).

\subsection{Phase 2: Interventional Factor Optimization}
\label{sec:method:phase2}
\subsubsection{Interventional Single-Factor Updates}
\label{sec:method:factorwise}

\paragraph{Interventional factor-wise updates.}
aPSF performs coordinate-style updates by performing a single-factor edit while freezing the others, and evaluates candidates via controlled substitutions.

Formally, let $B^{(t)}$ denote the factor contents at iteration $t$, and $p^{(t)} = \mathrm{Assemble}(B^{(t)})$ be the current prompt.
To optimize factor $\mathcal{F}_k$, the \textbf{Architect LLM $M_A$} proposes $N$ candidate edits $\mathcal{C}_k^{(t)}=\{c_k^{(j)}\}_{j=1}^{N}$.
We isolate each candidate's context-conditioned marginal contribution via contrastive evaluation. Let $B^{(t)}[k \leftarrow c]$ replace factor $k$ by $c$:
\begin{align}
\label{eq:intervene}
p^{(t)}_{k \leftarrow c} &= \mathrm{Assemble}(B^{(t)}[k \leftarrow c]), \\
\label{eq:score}
\mathrm{Score}(c) &= \hat{S}(p^{(t)}_{k \leftarrow c}, \mathcal{D}_{\mathrm{val}}),
\end{align}
where all other factors are held fixed. Let $\Delta \hat{S}(c)=\mathrm{Score}(c)-\hat{S}(p^{(t)},\mathcal{D}_{\mathrm{val}})$ denote the validation-score gain of candidate $c$, and select the best candidate edit $\hat{c}_k=\argmax_{c\in\mathcal{C}_k^{(t)}} \Delta \hat{S}(c)$. We accept the update if $\Delta \hat{S}(\hat{c}_k)\ge \delta$ and run for at most $T_{\max}$ steps, returning the best-performing prompt on $\mathcal{D}_{\mathrm{val}}$.
The complete procedure is detailed in \textbf{Algorithm~\ref{alg:phase2}} (Appendix~\ref{app:algorithms}).

\subsubsection{Error-Guided Factor Selection}
\label{sec:method:selection}

After a warm-start pass that optimizes each factor once, at each step $t$ we evaluate the current prompt on $\mathcal{D}_{\mathrm{val}}$ and collect representative failure cases.
We then summarize errors into a lightweight error profile and provide it to the Architect $M_A$ to select the bottleneck factor for the next update (e.g., \textsc{Rationale} for reasoning failures, \textsc{Format} for formatting errors).
Concretely, we use $M_A$ with a fixed, task-agnostic meta-prompt to produce \emph{open-ended} diagnoses for each failure (Appendix~\ref{app:metaprompts:diagnose}); for reporting we summarize the diagnoses into a lightweight error profile, while factor selection conditions on representative samples and the full diagnosis.
This error-guided routing makes the optimization \emph{adaptive}: the update budget is dynamically reallocated to whichever factor is most responsible for current failures.
If the same factor is repeatedly selected without improvement, $M_A$ may explore alternative factors to avoid local optima.
We stop early once $\hat{S}(p^{(t)}, \mathcal{D}_{\mathrm{val}})=1$; Figure~\ref{fig:execution_trace} (Appendix) provides an execution trace.

\begin{table*}[t]
\centering
\small
\setlength{\tabcolsep}{2pt}
\renewcommand{\arraystretch}{1.3}
\newcommand{\sm}[1]{{\tiny$\pm$#1}}
\newcommand{\n}[1]{{\footnotesize #1}}

\begin{tabular}{l||cccc|c||cccc|c}
\toprule
\multirow{2}{*}{\textbf{Methods}} 
& \multicolumn{5}{c||}{\textbf{Qwen2.5-7B-Instruct (Worker)}} 
& \multicolumn{5}{c}{\textbf{Llama-3.1-8B-Instruct (Worker)}} \\
\cmidrule(lr){2-6} \cmidrule(lr){7-11}
& GSM8K & AQUA & Multi & Hard & \textit{Avg}
& GSM8K & AQUA & Multi & Hard & \textit{Avg} \\
\midrule
\makecell{\textbf{Zero-shot}\\\textbf{CoT}} & \n{86.22}\sm{0.00} & \n{81.50}\sm{0.00} & \n{96.46}\sm{0.00} & \n{53.79}\sm{0.00} & \n{79.49}
              & \n{82.81}\sm{0.00} & \n{68.00}\sm{0.00} & \n{95.21}\sm{0.00} & \n{33.60}\sm{0.00} & \n{69.91} \\
\midrule
\multicolumn{11}{l}{\textit{Architect = Qwen3-8B}} \\
\quad OPRO & \n{87.63}\sm{1.85} & \n{77.00}\sm{3.42} & \n{98.75}\sm{0.38} & \n{53.60}\sm{2.91} & \n{79.25}
           & \n{84.95}\sm{1.67} & \n{64.50}\sm{4.23} & \n{96.46}\sm{0.52} & \n{33.62}\sm{3.58} & \n{69.88} \\
\quad APE & \n{88.83}\sm{0.92} & \underline{\n{81.50}}\sm{2.76} & \n{98.54}\sm{0.61} & \n{51.70}\sm{3.84} & \n{80.14}
          & \n{83.48}\sm{2.13} & \n{66.00}\sm{3.15} & \n{96.67}\sm{0.73} & \n{33.05}\sm{4.12} & \n{69.80} \\
\quad ProTeGi & \n{88.43}\sm{1.53} & \n{79.50}\sm{3.89} & \n{98.54}\sm{0.44} & \n{53.88}\sm{2.36} & \n{80.09}
              & \n{81.07}\sm{2.78} & \n{62.50}\sm{4.51} & \n{96.46}\sm{0.68} & \n{33.52}\sm{3.27} & \n{68.39} \\
\quad GrIPS & \n{88.16}\sm{2.14} & \n{80.50}\sm{2.93} & \n{98.54}\sm{0.57} & \n{52.56}\sm{3.45} & \n{79.94}
            & \n{83.55}\sm{1.89} & \underline{\n{69.00}}\sm{3.67} & \n{96.67}\sm{0.49} & \n{32.86}\sm{4.38} & \n{70.52} \\
\quad ZERA & \underline{\n{89.98}}\sm{0.63} & \n{81.36}\sm{0.75} & \n{98.42}\sm{0.43} & \n{53.22}\sm{0.94} & \n{80.75}
            & \n{84.15}\sm{0.80} & \n{65.95}\sm{0.95} & \underline{\n{98.22}}\sm{0.38} & \n{34.58}\sm{0.89} & \n{70.73} \\
\quad CriSPO & \n{89.40}\sm{0.69} & \n{81.48}\sm{0.97} & \underline{\n{98.94}}\sm{0.58} & \textbf{\n{55.94}}\sm{0.98} & \n{81.44}
             & \textbf{\n{89.49}}\sm{0.83} & \n{65.86}\sm{0.68} & \n{96.49}\sm{0.65} & \underline{\n{34.94}}\sm{0.51} & \n{71.70} \\
\rowcolor{lightblue}
\quad \textbf{aPSF} & \textbf{\n{90.03}}\sm{0.71} & \textbf{\n{82.50}}\sm{1.83} & \textbf{\n{99.30}}\sm{0.32} & \underline{\n{54.77}}\sm{1.94} & \textbf{\n{81.65}}$_{\textcolor{darkgreen}{\scriptscriptstyle +0.21}}$
                   & \underline{\n{85.62}}\sm{0.85} & \textbf{\n{71.50}}\sm{2.36} & \textbf{\n{98.33}}\sm{0.41} & \textbf{\n{35.98}}\sm{2.47} & \textbf{\n{72.86}}$_{\textcolor{darkgreen}{\scriptscriptstyle +1.16}}$ \\

\midrule
\multicolumn{11}{l}{\textit{Architect = gpt-oss-120b}} \\
\quad OPRO & \n{88.70}\sm{1.42} & \n{78.50}\sm{3.68} & \n{97.29}\sm{0.83} & \n{51.69}\sm{3.21} & \n{79.05}
           & \n{85.25}\sm{1.56} & \n{67.50}\sm{3.94} & \n{95.21}\sm{0.91} & \n{34.47}\sm{3.82} & \n{70.61} \\
\quad APE & \n{87.36}\sm{2.31} & \n{80.00}\sm{2.84} & \n{86.46}\sm{2.15} & \n{54.55}\sm{2.67} & \n{77.09}
          & \n{80.20}\sm{2.45} & \n{64.00}\sm{4.27} & \n{96.88}\sm{0.64} & \n{31.07}\sm{4.53} & \n{68.04} \\
\quad ProTeGi & \n{89.70}\sm{1.18} & \n{80.50}\sm{3.25} & \n{98.12}\sm{0.56} & \underline{\n{55.59}}\sm{2.83} & \n{80.98}
              & \n{84.48}\sm{1.73} & \n{53.00}\sm{4.86} & \n{97.50}\sm{0.52} & \underline{\n{35.88}}\sm{3.14} & \n{67.72} \\
\quad GrIPS & \underline{\n{90.80}}\sm{1.76} & \underline{\n{82.00}}\sm{2.51} & \n{98.54}\sm{0.48} & \n{53.41}\sm{3.57} & \n{81.19}
            & \n{84.28}\sm{2.08} & \n{68.00}\sm{3.42} & \n{96.88}\sm{0.71} & \n{32.95}\sm{3.96} & \n{70.53} \\
\quad ZERA & \n{89.55}\sm{0.60} & \n{80.31}\sm{0.75} & \textbf{\n{99.59}}\sm{0.89} & \n{52.63}\sm{0.85} & \n{80.52}
            & \underline{\n{89.69}}\sm{0.21} & \underline{\n{68.43}}\sm{0.51} & \textbf{\n{98.15}}\sm{0.78} & \n{34.70}\sm{0.84} & \n{72.74} \\
\quad CriSPO & \n{90.11}\sm{0.73} & \n{80.99}\sm{0.65} & \underline{\n{99.35}}\sm{0.51} & \n{51.25}\sm{0.53} & \n{80.43}
             & \textbf{\n{90.13}}\sm{0.89} & \n{65.13}\sm{0.84} & \n{97.90}\sm{1.10} & \n{35.16}\sm{1.15} & \n{72.08} \\
\rowcolor{lightblue}
\quad \textbf{aPSF} & \textbf{\n{90.87}}\sm{0.63} & \textbf{\n{83.00}}\sm{1.52} & \underline{\n{99.53}}\sm{0.29} & \textbf{\n{55.86}}\sm{1.68} & \textbf{\n{82.32}}$_{\textcolor{darkgreen}{\scriptscriptstyle +1.13}}$
                   & \n{86.22}\sm{0.78} & \textbf{\n{72.50}}\sm{2.18} & \underline{\n{97.92}}\sm{0.36} & \textbf{\n{42.96}}\sm{2.31} & \textbf{\n{74.90}}$_{\textcolor{darkgreen}{\scriptscriptstyle +2.16}}$ \\

\bottomrule
\end{tabular}

\caption{Test accuracy (\%) on math reasoning benchmarks (mean$_{\pm\text{std}}$ over 5 runs; ). Multi = MultiArith; Hard = GSM-Hard. All methods start from the same seed prompt (``Let's think step by step''). \textbf{Bold} = best; \underline{underline} = second best. Light blue rows highlight aPSF with gains in \textcolor{darkgreen}{green}.}
\label{tab:math_main_grouped}
\end{table*}

\section{Experiments}
\label{sec:setup}

\paragraph{Experimental questions.}
In this section, we conduct experiments to address the following questions:
\begin{itemize}[leftmargin=*, itemsep=2pt, topsep=2pt]
\item \textbf{Q1:} Can aPSF achieve competitive performance compared to strong API-only prompt optimizers under identical experimental conditions? See Section~\ref{sec:results}.
\item \textbf{Q2:} Does aPSF provide interpretable optimization behavior and component-level insights? See Section~\ref{sec:factor_patterns} and Section~\ref{sec:ablations}.
\item \textbf{Q3:} Does aPSF improve optimization efficiency, measured by total optimization tokens and steps to reach peak validation accuracy? See Section~\ref{sec:cost_analysis}.
\item \textbf{Q4:} Can aPSF generalize across tasks and transfer prompts across related datasets? See Section~\ref{sec:transfer}.
\end{itemize}

\subsection{Experimental Setup and Protocol}
\label{sec:setup:exp_setup}

\paragraph{Benchmarks and baselines.}
We evaluate on GSM8K~\citep{cobbe2021training}, AQUA-RAT~\citep{ling2017program}, MultiArith~\citep{roy2015solving}, GSM-Hard~\citep{gao2022pal}, BBH~\citep{srivastava2023beyond,suzgun2023challenging} (17 representative sub-tasks), and MMLU~\citep{hendrycks2020measuring}. We compare against state-of-the-art API-only prompt optimizers: OPRO~\citep{yang2023large}, ProTeGi~\citep{pryzant2023automatic}, APE~\citep{zhou2022large}, GrIPS~\citep{prasad2023grips}, ZERA~\citep{yi2025zera}, CriSPO~\citep{he2025crispo}, and the unoptimized initial prompt baseline (Zero-shot CoT, i.e., using $p_{\mathrm{init}}$ without optimization). All methods share the same Worker model, optimization budget, and fixed 50-example validation slice, and start from the same initial prompt $p_{\mathrm{init}}$; for aPSF, the initial factorization step is included in $T_{\max}$. We report final metrics on held-out test sets. Additional results on Competition Mathematics~\citep{hendrycks2021measuring} and GPQA~\citep{rein2024gpqa} are in Appendix~\ref{app:gpqa_math_results}.

\begin{figure*}[t]
  \centering
  \includegraphics[width=\linewidth]{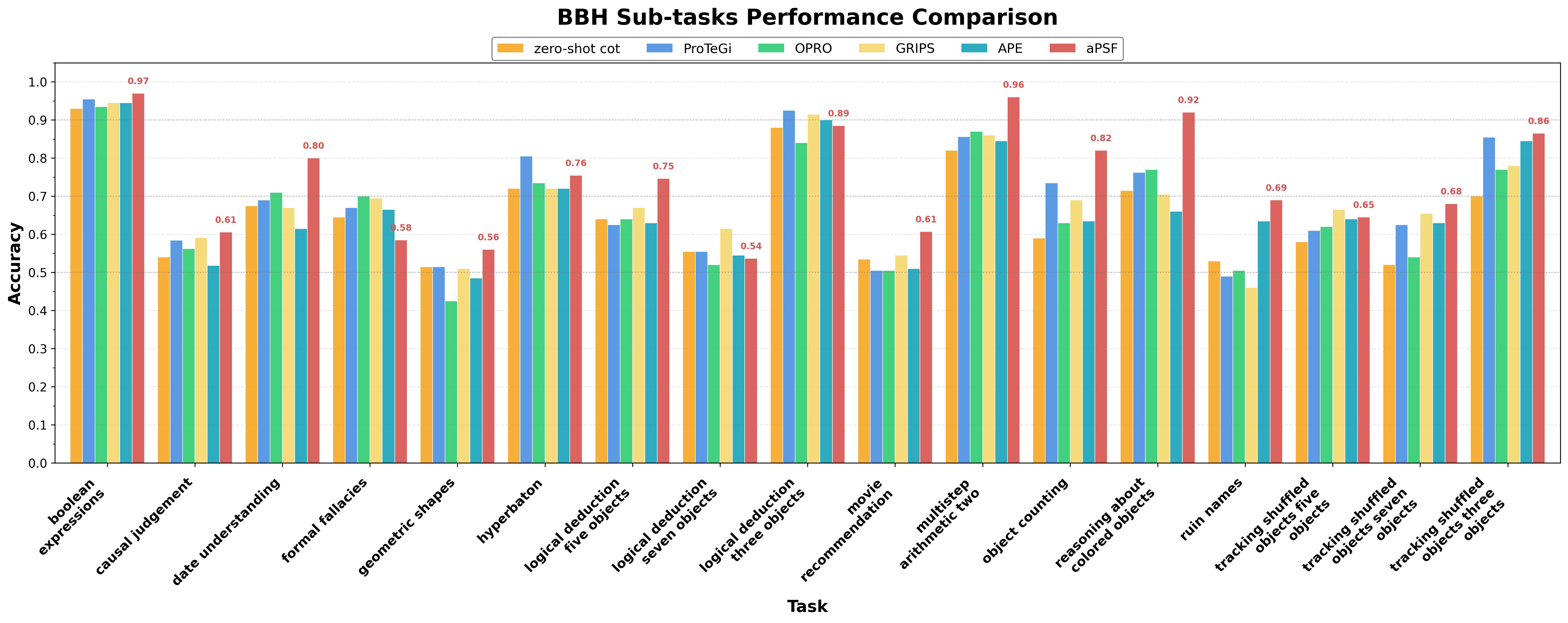}
  \caption{Per-task accuracy on BBH benchmarks. aPSF (red) achieves dominant performance on arithmetic and symbolic reasoning tasks, such as \textit{Multistep Arithmetic} (0.96) and \textit{Boolean Expressions} (0.97).}
  \label{fig:bbh_subtasks_algorithm}
\end{figure*}

\paragraph{Models and hyperparameters.}
We use \textbf{Qwen2.5-7B-Instruct}~\citep{team2024qwen2} as the default Worker LLM and \textbf{Qwen3-8B}~\citep{yang2025qwen3} as the Architect LLM. We additionally evaluate with \textbf{Llama-3.1-8B-Instruct}~\citep{dubey2024llama} as Worker and \textbf{gpt-oss-120b}~\citep{openai2025gptoss120bgptoss20bmodel} as Architect to test generalization. Architect quality can affect optimization; we report results with multiple Architects (Table~\ref{tab:math_main_grouped}). Throughout, we restrict ourselves to query-only access (prompt in, text out), without using model internals (e.g., gradients or logits). We run the Worker with temperature=0 for deterministic scoring and set the Architect temperature to 0.7 to encourage diverse candidate generation. Unless otherwise specified, we use $N{=}4$ candidates per update (Appendix~\ref{app:hyperparam}), an acceptance threshold $\delta{=}1/|\mathcal{D}_{\mathrm{val}}|$ (requiring at least one additional correct validation example), and cap optimization at $T_{\max}{=}10$ steps. We also instruct the Architect to prefer concise factorizations to avoid over-fragmentation. The factorization is induced by the Architect per task (Section~\ref{sec:method:phase1}). Due to stochasticity in candidate generation, we run the full optimization pipeline 5 times and report average test accuracy; all methods use the same protocol. Complete configurations are provided in Appendix~\ref{app:model_config}.

\subsection{Q1: aPSF Achieves the Best Accuracy Under Matched Budgets}
\label{sec:results}

We evaluate aPSF under matched API-only settings on (i) mathematical reasoning benchmarks (Table~\ref{tab:math_main_grouped}), (ii) diverse BBH reasoning sub-tasks (Figure~\ref{fig:bbh_subtasks_algorithm}), and (iii) a budget-matched comparison to programmatic prompting (DSPy; Table~\ref{tab:dspy_comparison}). Additional results on MMLU are reported in Appendix~\ref{app:mmlu_results}.
As shown in Table~\ref{tab:math_main_grouped}, Figure~\ref{fig:bbh_subtasks_algorithm}, and Table~\ref{tab:dspy_comparison}, we observe that:

\begin{table}[h]
\centering
\small
\begin{tabular}{lcccc}
\toprule
\textbf{Method} & \textbf{GSM8K} & \textbf{AQUA} & \textbf{GSM-Hard} & \textbf{MATH} \\
\midrule
DSPy & 89.83 & 81.00 & 51.59 & 52.80 \\
aPSF & \textbf{90.03} & \textbf{82.50} & \textbf{54.77} & \textbf{56.00} \\
\midrule
$\Delta$ & \textcolor{darkgreen}{\textbf{+0.20}} & \textcolor{darkgreen}{\textbf{+1.50}} & \textcolor{darkgreen}{\textbf{+3.18}} & \textcolor{darkgreen}{\textbf{+3.20}} \\
\bottomrule
\end{tabular}
\caption{Comparison with DSPy under matched settings. MATH = Competition Mathematics~\citep{hendrycks2021measuring}.}
\label{tab:dspy_comparison}
\end{table}

\begin{enumerate}[leftmargin=*, itemsep=1pt, topsep=1pt]
\item \textbf{Matched-performance:} aPSF achieves the best \textit{Avg} in all Worker/Architect settings on different benchmarks (Table~\ref{tab:math_main_grouped}), improving over the strongest baseline (including principle-aware methods CriSPO and ZERA) by up to \textbf{+2.16}pp and reaching \textbf{74.90\%} (vs\ 72.74\%) in the strongest setting.
\item \textbf{Broad coverage:} on BBH, aPSF shows consistent improvements across diverse reasoning patterns, with particularly strong margins on long-context, state-tracking tasks.
\item \textbf{Programmatic baselines:} under a controlled, budget-matched setup, aPSF outperforms DSPy across all datasets reported, suggesting advantages from task-adaptive factorization and error-guided updates beyond fixed program scaffolds.
\end{enumerate}

\subsubsection{Q1-BBH: Complex Reasoning}
\label{sec:logical_reasoning}

We evaluate aPSF on 17 representative BBH sub-tasks covering arithmetic, symbolic, logical, and temporal reasoning. Figure~\ref{fig:bbh_subtasks_algorithm} reports per-task accuracies against baselines.

aPSF delivers the largest gains on \textbf{Multistep Arithmetic Two} (+9.0pp vs OPRO) and \textbf{Reasoning about Colored Objects} (+15.0pp), consistent with improved long-context state tracking from factorizing the prompt into separable components (e.g., step-wise reasoning and consistency checks). Beyond these extremes, aPSF achieves high absolute accuracy on symbolic and state-tracking tasks such as \textbf{Boolean Expressions} (0.97), \textbf{Web of Lies} (0.905), and \textbf{Tracking Shuffled Objects Three} (0.86). It also improves logical deduction (e.g., +4.5pp on \textbf{Logical Deduction Three Objects}). This robustness is further reflected in the performance profile: within 10\% of the best method ($\tau{=}0.1$), aPSF succeeds on 76\% of BBH tasks vs.\ 65\% for OPRO (Figure~\ref{fig:performance_profile}).

\subsubsection{Q1-DSPy: Comparison to Programmatic Prompting}
\label{sec:results:dspy}

\begin{figure*}[t]
\centering
\includegraphics[width=\textwidth]{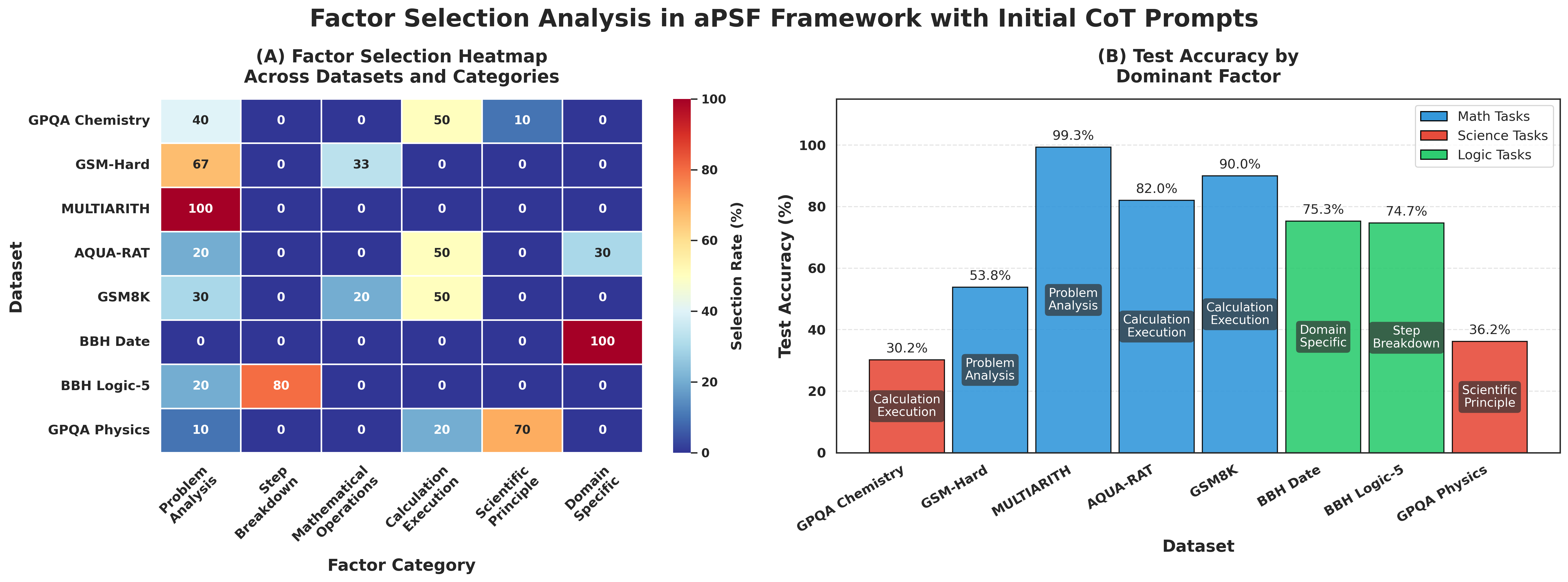}
\caption{
    \textbf{Factor selection patterns and performance.}
    (\textit{A})~Selection rate heatmap grouped by semantic category (original names in Appendix~\ref{app:factor_names}).
    (\textit{B})~Test accuracy and dominant factors per task (color indicates domain).
}
\label{fig:factor_patterns}
\end{figure*}

We compare aPSF with DSPy~\citep{khattab2024dspy}, a programmatic prompting framework that compiles prompts and demonstrations for a fixed, user-specified program scaffold.
For a budget-matched setup, we use a single \texttt{ChainOfThought} module and compile it with \texttt{BootstrapFewShotWithRandomSearch}, running 10 trials ($T_{\max}{=}10$) on the same validation slice, without finetuning or external data.
Both methods use the same Worker (Qwen2.5-7B-Instruct) and the same compiler/optimizer LLM (Qwen3-8B) for prompt proposal/compilation.
As shown in Table~\ref{tab:dspy_comparison}, aPSF outperforms DSPy on all reported datasets, which we attribute to task-adaptive factorization and error-guided updates rather than DSPy’s fixed single-module scaffold.
Programmatic systems (e.g., SAMMO) and task-faceting methods~\citep{juneja2025task} are discussed in Related Work but are not evaluated, as their interfaces and scaffold assumptions conflict with our query-only setting.

\begin{table}[h]
\centering
\small
\renewcommand{\arraystretch}{1.15}
\setlength{\tabcolsep}{5pt}
\resizebox{\columnwidth}{!}{%
\begin{tabular}{lccc}
\toprule
\textbf{Task} & \textbf{aPSF} & \textbf{NoStructure} & \textbf{Gain ($\uparrow$)} \\
\midrule
GSM8K & 90.03 & 89.55 & \textcolor{darkgreen}{\textbf{+0.48}} \\
AQUA-RAT & 82.50 & 79.50 & \textcolor{darkgreen}{\textbf{+3.00}} \\
Date Understanding & 80.00 & 74.00 & \textcolor{darkgreen}{\textbf{+6.00}} \\
Web of Lies & 90.50 & 85.30 & \textcolor{darkgreen}{\textbf{+5.20}} \\
Movie Recommendation & 60.72 & 58.67 & \textcolor{darkgreen}{\textbf{+2.05}} \\
\midrule
\textbf{Average} & 80.75 & 77.40 & \textcolor{darkgreen}{\textbf{+3.35}} \\
\bottomrule
\end{tabular}}
\caption{System-level ablation. Task-specific factorization is critical for logical reasoning tasks (e.g., Date Understanding), yielding significant gains (+6.00pp) over fixed templates.}
\label{tab:bbh_nostructure}
\end{table}

\subsection{Q2: Routing Is Interpretable and Structure Matters}
\label{sec:ablations}

\begin{figure}[t]
\centering
\includegraphics[width=1\linewidth]{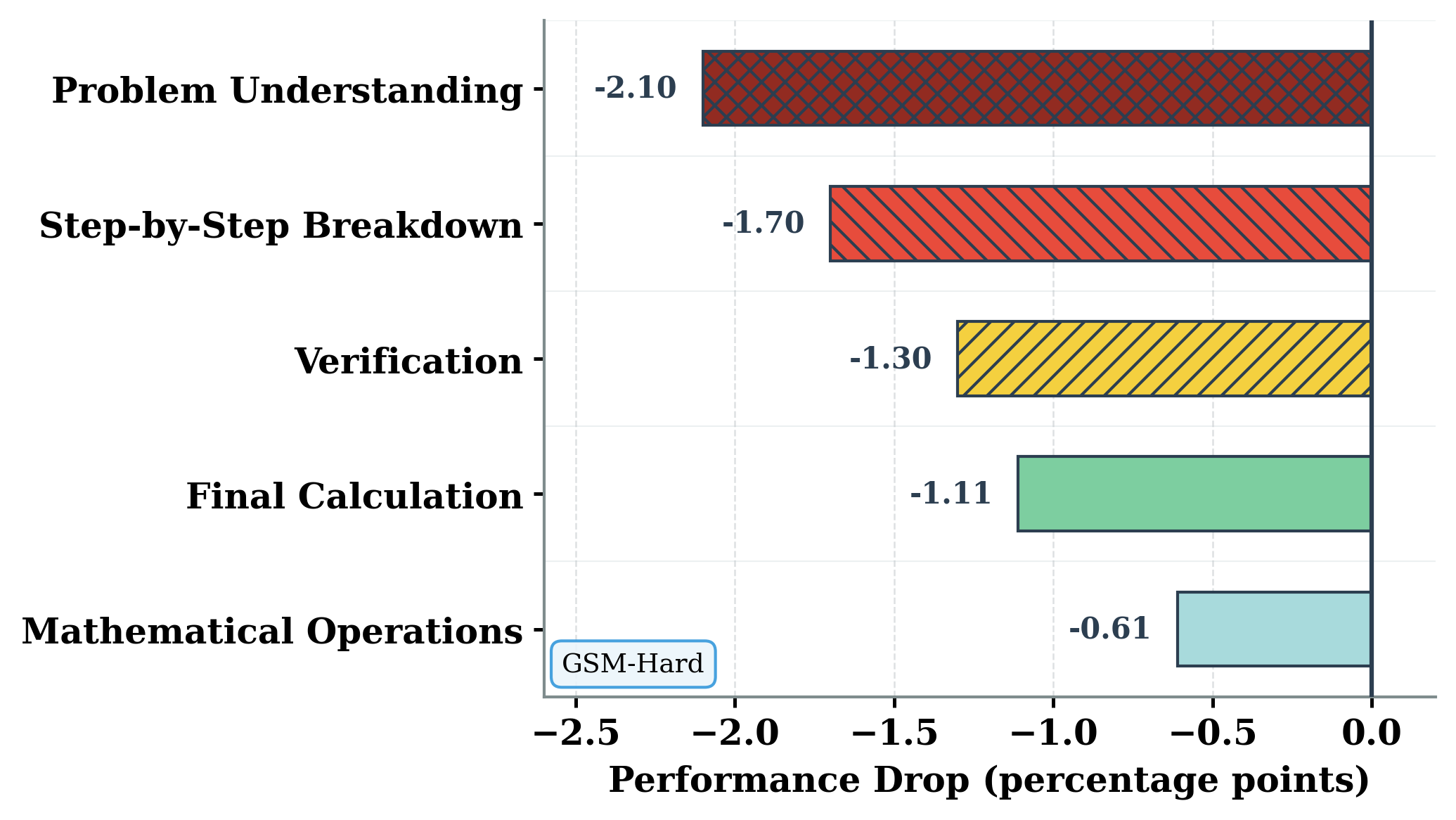}
\caption{Factor-level ablation on GSM-Hard. Bars show performance drop when each factor is removed.}
\label{fig:removed_factor}
\end{figure}

\noindent We analyze optimization behavior via routing patterns across tasks and controlled ablations of structure, scheduling, and factor freezing (Figure~\ref{fig:factor_patterns}, Table~\ref{tab:bbh_nostructure}, Figure~\ref{fig:removed_factor}, Table~\ref{tab:scheduler_gsm8k_aqua_ranked}, Table~\ref{tab:frozen_ablation}).
As shown, we observe that:
\begin{enumerate}[leftmargin=*, itemsep=1pt, topsep=1pt]
\item \textbf{Interpretable routing:} error-guided factor selection yields task-adaptive factor usage aligned with domain needs (e.g., math emphasizes \textsc{Problem Analysis}/\textsc{Calculation Execution}; science emphasizes \textsc{Scientific Principle}), and the dominant factors correlate with strong per-task accuracy (Figure~\ref{fig:factor_patterns}).
\item \textbf{Structure matters:} removing task-specific factorization (\textbf{NoStructure}) causes a sizable average drop (3.35pp) and disproportionately hurts format-sensitive BBH tasks, indicating generic templates miss critical procedural structure (Table~\ref{tab:bbh_nostructure}).

\begin{table}[h]
\centering
\small
\renewcommand{\arraystretch}{1.15}
\setlength{\tabcolsep}{5pt}
\begin{tabular}{lSSS}
\toprule
\textbf{Method} & {\textbf{GSM8K}} & {\textbf{AQUA}} & {\textbf{Average}} \\
\midrule
\rowcolor{lightblue}
\textbf{Error-Guided} & 90.03 & \bfseries 82.50 & \bfseries 86.27$_{\textcolor{darkgreen}{\scriptscriptstyle +1.53}}$ \\
Thompson             & \bfseries 90.57 & 78.00 & 84.29  \\
Round-robin          & 90.03 & 79.00 & 84.52  \\
Greedy               & 89.97 & 79.50 & 84.74 \\
Random               & 89.80 & 79.50 & 84.65 \\
\bottomrule
\end{tabular}
\caption{Scheduler ablation on GSM8K and AQUA (Accuracy \%). Error-Guided achieves the best average, while Thompson sampling shows higher variance across datasets.}
\label{tab:scheduler_gsm8k_aqua_ranked}
\end{table}

\item \textbf{Actionable components:} factor-level knock-outs identify which components are bottlenecks (Figure~\ref{fig:removed_factor}), and scheduler ablations show error-guided factor selection improves stability and average accuracy versus alternative update strategies (Table~\ref{tab:scheduler_gsm8k_aqua_ranked}).
\item \textbf{Single-factor updates:} unfrozen multi-factor edits degrade accuracy by an average of 1.03pp, confirming that freezing non-target factors improves credit assignment and optimization stability (Table~\ref{tab:frozen_ablation}).
\end{enumerate}

\subsubsection{Q2-Routing: Adaptive Patterns Across Tasks}
\label{sec:factor_patterns}

We analyze how error-guided factor selection allocates optimization effort across tasks (Figure~\ref{fig:factor_patterns}).

\paragraph{Task-adaptive optimization.}
Figure~\ref{fig:factor_patterns} shows task-specific routing patterns: math concentrates on \textsc{Problem Analysis}/\textsc{Calculation Execution}, logic relies on \textsc{Step Breakdown} or \textsc{Domain Specific} factors, and science emphasizes \textsc{Scientific Principle}. Dominant factors align with stronger test accuracy, indicating domain-specific factor priorities.
Discovered schemas and trajectories are in Appendix~\ref{app:factor_names} (Tables~\ref{tab:factor_names_detailed} and~\ref{tab:factor_examples}) and Appendix~\ref{app:trajectory} (Figures~\ref{fig:bbh_date_stepwise} and~\ref{fig:execution_trace}).

\subsubsection{Q2-Ablations: Component Knock-Outs}
\label{sec:ablations:knockouts}

We analyze component importance via system-level structure and factor-level sensitivity.

\begin{table}[h]
  \centering
  \small
  \resizebox{\columnwidth}{!}{
  \begin{tabular}{l|ccccc}
  \toprule
  \multirow{2}{*}{\textbf{Source}} & \multicolumn{5}{c}{\textbf{Target Dataset}} \\
  \cmidrule(lr){2-6}
   & GSM8K & AQUA & MultiArith & GSM-Hard & MATH \\
  \midrule
  MultiArith & 89.76 & 81.50 & -- & 52.98 & 52.80 \\
  GSM8K      & -- & 82.00 & 99.30 & 54.08 & 51.60 \\
  GSM-Hard   & 89.27 & 80.50 & 99.30 & -- & 51.00 \\
  AQUA       & 90.31 & -- & 99.30 & 53.78 & 52.20 \\
  \midrule
  \rowcolor{gray!15}
  \textit{Same-task} & \textit{90.03} & \textit{82.50} & \textit{99.30} & \textit{54.77} & \textit{56.00} \\
  \bottomrule
  \end{tabular}
  }
  \caption{
\textbf{Cross-task transfer results.}
Prompts optimized on source tasks, evaluated on targets.
Last row: same-task baseline. MATH = Competition Mathematics~\citep{hendrycks2021measuring}.
}

  \label{tab:transfer}
\end{table}

\paragraph{System-level: Impact of Structure.}
We compare against \textbf{NoStructure}, a fixed template without task-specific factorization. As shown in Table~\ref{tab:bbh_nostructure}, accuracy drops by 3.35pp on average: the effect is small on GSM8K ($-0.48$pp) but larger on format-sensitive BBH tasks (e.g., \textit{Date Understanding}: 80.0$\to$74.0, $-6.0$pp), indicating generic templates miss procedural structure.

\paragraph{Factor-level: Fine-grained Sensitivity.}
Fig.~\ref{fig:removed_factor} masks individual factors on GSM-Hard. Removing \emph{Problem Understanding} causes the largest drop ($-2.10$pp), while \emph{Mathematical Operations} has a smaller effect ($-0.61$pp), suggesting the Worker handles much of the arithmetic internally.

\subsubsection{Q2-Scheduler: Ablations}

We compare error-guided factor selection with Round-robin, Thompson Sampling, and Greedy under identical settings (Table~\ref{tab:scheduler_gsm8k_aqua_ranked}). Error-Guided achieves the best average across GSM8K and AQUA (86.27\%), while Thompson sampling attains the highest GSM8K score but degrades on AQUA, indicating higher variance. This supports the claim that adaptive factor selection based on step-level error analysis improves stability without sacrificing overall quality.

\subsubsection{Q2-Freezing: Single-Factor vs.\ Multi-Factor Updates}
\label{sec:ablations:freezing}

To validate the single-factor update mechanism, we compare aPSF's default frozen single-factor updates against an \textbf{unfrozen} variant that allows edits to modify multiple factors per step. Under the same budget ($T_{\max}{=}10$), Worker/Architect setup, and evaluation protocol, the unfrozen variant consistently underperforms (Table~\ref{tab:frozen_ablation}).

\begin{table}[h]
\centering
\small
\renewcommand{\arraystretch}{1.15}
\setlength{\tabcolsep}{4pt}
\begin{tabular}{lccc}
\toprule
\textbf{Dataset} & \textbf{Frozen} & \textbf{Unfrozen} & \textbf{Gain ($\uparrow$)} \\
\midrule
AQUA-RAT & \textbf{82.50} & 81.50 & \textcolor{darkgreen}{\textbf{+1.00}} \\
MultiArith & \textbf{99.30} & 97.08 & \textcolor{darkgreen}{\textbf{+2.22}} \\
GSM-Hard & \textbf{54.77} & 54.36 & \textcolor{darkgreen}{\textbf{+0.41}} \\
Date Understanding & \textbf{80.00} & 79.50 & \textcolor{darkgreen}{\textbf{+0.50}} \\
\midrule
\textbf{Average} & \textbf{79.14} & 78.11 & \textcolor{darkgreen}{\textbf{+1.03}} \\
\bottomrule
\end{tabular}
\caption{Frozen (single-factor) vs.\ unfrozen (multi-factor) updates. Freezing non-target factors yields higher accuracy by preventing credit-assignment noise from coupled edits.}
\label{tab:frozen_ablation}
\end{table}

When factors are unfrozen, edits spill over to adjacent factors, confounding credit assignment and producing noisier trajectories. This confirms that single-factor interventions improve stability and attribution, corroborating the coordinate-descent design in Section~\ref{sec:method:factorwise}.

\subsection{Q3: aPSF Reaches Peak Validation in 1 Step with 45--87\% Fewer Tokens}
\label{sec:cost_analysis}

\begin{figure}[t]
\centering
\includegraphics[width=1\linewidth]{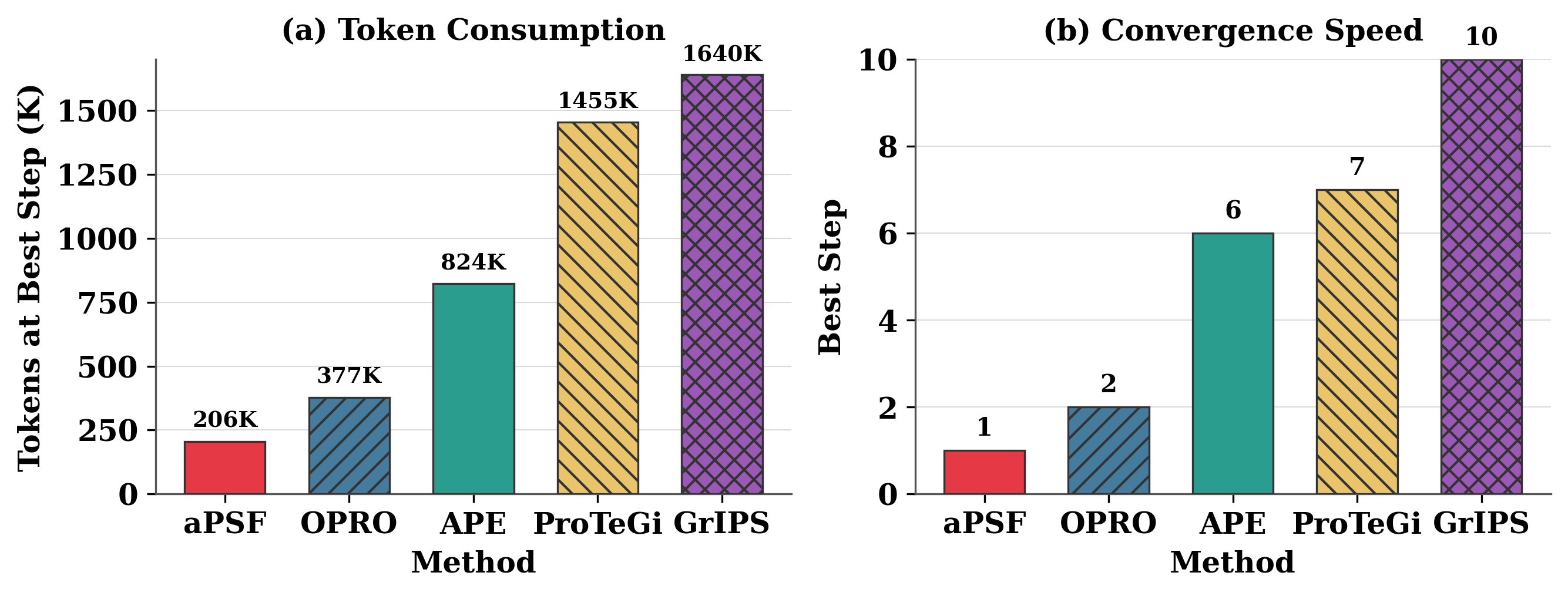}
\caption{Optimization efficiency on MultiArith. (a) Token consumption; (b) Convergence steps (lower is better).}
\label{fig:token_efficiency}
\end{figure}

We analyze efficiency on MultiArith (Figure~\ref{fig:token_efficiency}) along two axes: \emph{total optimization tokens} and \emph{steps-to-best} (the number of optimization iterations until the run attains its maximum validation accuracy).
Total optimization tokens include both Architect generation (structure discovery and candidate proposals) and Worker evaluation (validation inference) across all steps.
As shown in Figure~\ref{fig:token_efficiency}, we observe that:
\begin{enumerate}[leftmargin=*, itemsep=1pt, topsep=1pt]
\item \textbf{Token efficiency:} aPSF reaches peak validation accuracy with 206K tokens (45--87\% fewer than baselines).
\item \textbf{Convergence speed:} aPSF reaches peak validation accuracy at step 1 (vs.\ 6--10 steps for others).
\item \textbf{Generalization:} This efficiency advantage generalizes across datasets; on GSM8K, AQUA, and GSM-Hard, aPSF consistently reaches peak validation earlier with 36--55\% fewer tokens than OPRO and ZERA (Appendix~\ref{app:efficiency_extended}).
\end{enumerate}

\subsection{Q4: Prompts Transfer Across Related Arithmetic Tasks}
\label{sec:transfer}

Table~\ref{tab:transfer} evaluates cross-task transfer by applying prompts optimized on a source dataset to other targets.
As shown in Table~\ref{tab:transfer}, we observe that:
\begin{enumerate}[leftmargin=*, itemsep=1pt, topsep=1pt]
\item Transfer is strong among related arithmetic tasks, sometimes matching or slightly exceeding same-task baselines.
\item More distant or harder targets exhibit larger drops, indicating task-specific bottlenecks remain.
\end{enumerate}

\section{Conclusion}
\label{sec:conclusion}

We introduce \textbf{Adaptive Prompt Structure Factorization (aPSF)}, which automatically discovers a task-specific prompt factorization and optimizes it via single-factor interventions and error-guided factor selection.
Across mathematical and logical reasoning benchmarks, aPSF delivers consistent gains and improves controllability and factor-level attribution over monolithic prompt editing.

\section{Limitations}
\label{sec:limitations}

Our framework has three main limitations.
First, the quality of structure discovery and error diagnosis relies on the capabilities of the Architect LLM; weaker models may induce suboptimal factorizations.
Second, our interventional factor-level scoring assumes partial independence between factors. While this isolates marginal contributions for clean credit assignment, it may underestimate complex synergies between coupled factors (as discussed in Section~\ref{sec:ablations:knockouts}).
Finally, our evaluation is currently limited to text-based reasoning benchmarks, leaving multimodal or tool-augmented scenarios for future exploration.

\section{Ethical Considerations}
\label{sec:ethics}

aPSF is a general prompt-optimization framework and could be misused to optimize prompts for harmful or policy-violating objectives, depending on the underlying model and usage context.
Our study focuses on established public reasoning benchmarks and does not collect new user data or involve human subjects.
In practice, responsible use should follow applicable platform policies and safety filters, and any released code or prompts should include usage guidelines that discourage harmful applications.

\bibliography{main}

\clearpage
\appendix

\section{Detailed Algorithms}
\label{app:algorithms}
Phase 2 algorithm follows.
\FloatBarrier

\begin{algorithm}[H]
\caption{Phase 2: Factor-wise Optimization (Error-Guided Factor Selection)}
\label{alg:phase2}
\begin{algorithmic}[1]
\State \textbf{Input:} Factor schema $\mathcal{G}$, initial content $B^{(0)}$, validation set $\mathcal{D}_{\mathrm{val}}$
\State \textbf{Hyperparams:} $N$, $T_{\max}$, $\delta$ (acceptance threshold), $P$ (patience)
\State Let $K \gets |\mathcal{G}|$; $\text{best\_B} \gets B^{(0)}$; $\text{best\_score} \gets \hat{S}(\mathrm{Assemble}(B^{(0)}),\mathcal{D}_{\mathrm{val}})$; $u \gets 0$. \Comment{$u$: no-improve counter}
\For{$t=0,1,\ldots,T_{\max}-1$}
  \If{$t<K$}
    \State $k \gets t+1$ \Comment{Warm-start: optimize each factor once sequentially}
  \Else
    \State Collect errors on $\mathcal{D}_{\mathrm{val}}$ using current prompt $p^{(t)} = \mathrm{Assemble}(B^{(t)})$
    \State Use $M_A$ to produce open-ended diagnoses for representative failures and summarize them into a lightweight error profile
    \State $k \gets M_A(p_{\mathrm{diagnose}}^{\mathrm{meta}}; \{\text{errors}\}, \mathcal{G}, p^{(t)})$ \Comment{Error-guided factor selection (conditions on diagnoses + samples)}
  \EndIf
  \State $\mathcal{C}_k^{(t)} \gets M_A(p_{\mathrm{edit}}^{\mathrm{meta}}; f_k^{(t)},\mathcal{G})$ \Comment{Architect proposes candidates}
  \For{each $c \in \mathcal{C}_k^{(t)}$}
    \State $p^{(t)}_{k \leftarrow c}\gets \mathrm{Assemble}(B^{(t)}[k \leftarrow c])$ \Comment{Substitute factor $k$}
    \State $\Delta \hat{S}(c) \gets \hat{S}(p^{(t)}_{k \leftarrow c},\mathcal{D}_{\mathrm{val}}) - \hat{S}(\mathrm{Assemble}(B^{(t)}),\mathcal{D}_{\mathrm{val}})$
  \EndFor
  \State $\hat{c}\gets \argmax_{c\in\mathcal{C}_k^{(t)}} \Delta \hat{S}(c)$;\quad $r\gets \Delta \hat{S}(\hat{c})$
  \If{$r\ge\delta$}
     \State $B^{(t{+}1)} \gets B^{(t)}[k \leftarrow \hat{c}]$ \Comment{Accept edit if gain $\ge\delta$}
  \Else
     \State $B^{(t{+}1)} \gets B^{(t)}$ \Comment{Keep current}
  \EndIf
  \State $s \gets \hat{S}(\mathrm{Assemble}(B^{(t{+}1)}),\mathcal{D}_{\mathrm{val}})$
  \State \textbf{if} $s>\text{best\_score}$: $\text{best\_score}\gets s$, $\text{best\_B}\gets B^{(t{+}1)}$, $u\gets 0$; \textbf{else} $u\gets u+1$
  \State \textbf{if} $s=1$ \textbf{or} $u\ge P$ \textbf{then break} \Comment{Perfect val or early stop}
\EndFor
\State \textbf{return} $\text{best\_B}$
\end{algorithmic}
\end{algorithm}

\section{Reproducibility Checklist}
\label{app:repro}

We provide comprehensive experimental details to ensure full reproducibility of our results:

\begin{itemize}[leftmargin=*, itemsep=3pt, topsep=4pt]
\item \textbf{Datasets:} All benchmark names, sources, and official split information are provided in Appendix~\ref{app:data}. Validation and test splits are fixed across all methods and random seeds. We provide exact indices used for validation/test partitioning.

\item \textbf{Model Configurations:} Complete model specifications are detailed in Appendix~\ref{app:model_config}. \textbf{Worker and Architect LLM} configurations remain identical across all experiments unless explicitly noted otherwise.

\item \textbf{Evaluation Protocol:} All methods use identical worker models, validation slice size (50 examples), decoding settings (see Appendix~\ref{app:model_config}), and candidate count $N{=}4$. Complete protocol details in Appendix~\ref{app:eval_protocol}.

\item \textbf{Meta-Prompts:} Full meta-prompts for structure discovery (both from-scratch and initial-prompt modes), factor-wise editing, open-ended error diagnosis, and factor selection are provided in Appendix~\ref{app:metaprompts}.

\item \textbf{Hyperparameters:} All hyperparameters with sensitivity analysis: candidates per update $N{=}4$, acceptance threshold $\delta{=}1/|\mathcal{D}_{\mathrm{val}}|$; maximum iterations $T_{\max}{=}10$. Sensitivity analysis in Appendix~\ref{app:hyperparam}.

\item \textbf{Random Seeds:} Due to Architect stochasticity (temperature=0.7), we run 5 independent optimization runs and report mean test accuracy. Worker inference is deterministic (temperature=0.0). Fixed seed 42 for data splits.

\item \textbf{Computational Resources:} Hardware and software stack specifications (GPU/CPU/RAM/OS) in Appendix~\ref{app:compute}.

\item \textbf{Use of AI Assistants:} We used an AI assistant solely for English proofreading and stylistic polishing. All technical content, experiments, and conclusions were produced and verified by the authors.

\item \textbf{Code and Artifacts Release:} Code and artifacts are publicly available at \url{https://github.com/luckyboyLiu/aPSF} under the MIT license, including: (1) complete source code with documentation, (2) data loaders for all benchmarks, (3) unified evaluation and scoring pipeline, (4) checkpoint-based resume support, and (5) configuration for multiple LLM backends (OpenAI-compatible, Ollama, vLLM).
\end{itemize}
\section{Meta-Prompts}
\label{app:metaprompts}

This section presents the complete meta-prompts used for structure discovery, factor-wise optimization, and error-guided factor selection. All meta-prompts are designed to be model-agnostic.

\subsection{Structure Discovery (From-Scratch Mode)}
\label{app:metaprompts:struct}

This meta-prompt instructs the Architect LLM to autonomously discover task-specific factor structures.

\begin{figure*}[h]
\centering
\begin{promptbox}[title={\textsc{From-Scratch Structure Discovery}}]
\small\ttfamily
Based on the task description and examples, generate a complete, usable instruction and decompose it into factors.

\vspace{2pt}
\textbf{[INPUT]}\\[2pt]
\hspace*{4pt}Task Description: \{task\_description\}\\
\hspace*{4pt}Task Type: \{task\_type\}\\
\hspace*{4pt}Expected Output: \{output\_format\}; constraints: \{constraints\}\\
\hspace*{4pt}Example Data: \{example\_data\}

\vspace{4pt}
Return ONLY the four sections below, plain text, no extra commentary:

\vspace{4pt}
\colorbox{blue!8}{\textbf{Section 1: Complexity Analysis}}\\[2pt]
\hspace*{4pt}[one sentence on why you chose K factors (prefer concise factorizations)]

\vspace{2pt}
\colorbox{blue!8}{\textbf{Section 2: Complete Instruction Template}}\\[2pt]
\hspace*{4pt}[concise, natural instruction; no numbering/bold/quotes]

\vspace{2pt}
\colorbox{blue!8}{\textbf{Section 3: Factor Decomposition}}\\[2pt]
\hspace*{4pt}Factor1\_[Name]: [one-line role]\\
\hspace*{4pt}Factor2\_[Name]: [one-line role]\\
\hspace*{4pt}... (use K factors; prefer fewer when possible)

\vspace{2pt}
\colorbox{blue!8}{\textbf{Section 4: Factor Boundary Mapping}}\\[2pt]
\hspace*{4pt}Factor1\_[Name]: "[verbatim substring from template]"\\
\hspace*{4pt}Factor2\_[Name]: "[verbatim substring from template]"\\
\hspace*{4pt}... (names must exactly match Factor Decomposition)

\vspace{4pt}
\textit{Do not output anything beyond these four sections.}
\end{promptbox}
\caption{Meta-prompt for \textbf{From-Scratch Structure Discovery}. It enforces a strict four-section output format to ensure reliable parsing of the factor structure.}
\label{fig:prompt_struct}
\end{figure*}

\subsection{Initial-Prompt Analysis and Extension}
\label{app:metaprompts:initial}

This meta-prompt enables aPSF to start from a user-provided initial prompt and systematically extend it.

\begin{figure*}[h]
\centering
\begin{promptbox}[title={\textsc{Meta-Prompt: Initial-Prompt Analysis}}]
\small\ttfamily
\textbf{[INPUT]}\\
Initial Prompt: \{initial\_prompt\}\\
Task Description: \{task\_description\}\\
Task Type: \{task\_type\}\\
Expected Output: \{output\_format\}; constraints: \{constraints\}\\
Example Data: \{example\_data\}

\vspace{4pt}
Based on the initial prompt and task examples, generate a complete, usable instruction and decompose it into factors. Return ONLY the four sections below:

\vspace{2pt}
\colorbox{blue!8}{\textbf{Section 1: Complexity Analysis}}\\[2pt]
\hspace*{4pt}[one sentence on why you chose K factors (prefer concise factorizations)]

\vspace{2pt}
\colorbox{blue!8}{\textbf{Section 2: Complete Instruction Template}}\\[2pt]
\hspace*{4pt}[concise, natural instruction; no numbering/bold/quotes]

\vspace{2pt}
\colorbox{blue!8}{\textbf{Section 3: Factor Decomposition}}\\[2pt]
\hspace*{4pt}Factor1\_[Name]: [one-line role]\\
\hspace*{4pt}Factor2\_[Name]: [one-line role]\\
\hspace*{4pt}... (use K factors; prefer fewer when possible)

\vspace{2pt}
\colorbox{blue!8}{\textbf{Section 4: Factor Boundary Mapping}}\\[2pt]
\hspace*{4pt}Factor1\_[Name]: ``[verbatim substring from template]''\\
\hspace*{4pt}Factor2\_[Name]: ``[verbatim substring from template]''\\
\hspace*{4pt}... (names must exactly match Factor Decomposition)

\vspace{4pt}
\textit{Do not output anything beyond these four sections.}
\end{promptbox}
\caption{Meta-prompt for \textbf{Initial-Prompt Analysis}. It guides the Architect to preserve the core reasoning style of the user's input while expanding it into a full factorized program.}
\label{fig:prompt_initial}
\end{figure*}

\subsection{Factor-Wise Edit Generation}
\label{app:metaprompts:edit}

This meta-prompt guides the Architect LLM to generate diverse candidate edits while avoiding overfitting to specific errors.

\begin{figure*}[h]
\centering
\begin{promptbox}[title={\textsc{Meta-Prompt: Factor-Wise Editing}}]
\small\ttfamily
You are optimizing a prompt for a \{dataset\_name\} dataset (\{task\_type\}).

\vspace{2pt}
\colorbox{blue!8}{\textbf{Current Performance}}\\[2pt]
\hspace*{4pt}Accuracy: \{accuracy\}\% (\{correct\_count\}/\{total\_samples\} correct)\\
\hspace*{4pt}The current prompt already works for \{correct\_count\} samples - DO NOT break what's working!

\vspace{2pt}
\colorbox{blue!8}{\textbf{Context}}\\[2pt]
\hspace*{4pt}Current Complete Prompt: \{current\_prompt\}\\
\hspace*{4pt}Target Factor Segment: ``\{current\_factor\_desc\}'' (This is the part to replace)\\
\hspace*{4pt}Role/Goal of this factor: \{target\_factor\}

\vspace{2pt}
\colorbox{blue!8}{\textbf{Error Analysis (for reference only)}}\\[2pt]
\hspace*{4pt}\{error\_summary\}

\vspace{2pt}
\colorbox{blue!8}{\textbf{Task}}\\[2pt]
\hspace*{4pt}Generate \{num\_candidates\} improved versions of the ``\{target\_factor\}'' segment.

\vspace{2pt}
\colorbox{red!15}{\textbf{CRITICAL CONSTRAINTS}}\\[2pt]
\hspace*{4pt}1. Output ONLY the new text segment.\\
\hspace*{4pt}2. The new segment must be grammatically compatible with the surrounding text.\\
\hspace*{4pt}3. PRESERVE what makes the current prompt work for correct samples.\\
\hspace*{4pt}4. Keep improvements CONCISE and GENERAL-PURPOSE for this dataset type.\\
\hspace*{4pt}5. Do NOT overfit to the specific error examples - improve the general approach.\\
\hspace*{4pt}6. Consider the nature of \{dataset\_name\} tasks when making improvements.\\
\hspace*{4pt}7. Do NOT include markdown blocks, just raw JSON.

\vspace{4pt}
\textit{Output format: A valid JSON array of strings, e.g., [``description 1'', ``description 2''].}
\end{promptbox}
\caption{Meta-prompt for \textbf{Factor-Wise Editing}. It provides error analysis as reference while explicitly preventing overfitting through general-purpose constraints.}
\label{fig:prompt_edit}
\end{figure*}

\begin{figure*}[h]
\centering
\begin{promptbox}[title={\textsc{Meta-Prompt: Step 1 -- Open-Ended Error Diagnosis}}]
\small\ttfamily
You are a professional AI error analysis expert. Your task is to conduct a thorough and unbiased diagnosis of an observed error in an AI-generated response.

\vspace{2pt}
\colorbox{blue!8}{\textbf{Input Information}}\\
\hspace*{4pt}Question: \{question\}\\
\hspace*{4pt}Correct Answer: \{correct\_answer\}\\
\hspace*{4pt}AI Predicted Answer: \{predicted\_answer\}\\
\hspace*{4pt}AI Reasoning Process: \{reasoning\}

\vspace{2pt}
\colorbox{blue!8}{\textbf{Diagnosis Objective}}\\
\hspace*{4pt}Perform an open-ended and comprehensive error analysis without being constrained\\
\hspace*{4pt}by predefined categories. Focus on understanding the underlying causes and implications.

\vspace{2pt}
\colorbox{blue!8}{\textbf{Analysis Dimensions}}\\
\hspace*{4pt}1. \textbf{Error Essence:} Identify the fundamental root cause of the error.\\
\hspace*{4pt}2. \textbf{Error Type:} Assign a concise and descriptive label to characterize the error.\\
\hspace*{4pt}3. \textbf{Error Mechanism:} Explain how and why the error occurred.\\
\hspace*{4pt}4. \textbf{Error Impact:} Assess how this error affects the overall reasoning or outcome.\\
\hspace*{4pt}5. \textbf{Improvement Direction:} Propose specific and actionable prompt-level improvements.

\vspace{2pt}
\colorbox{red!15}{\textbf{Response Format}}\\
\hspace*{4pt}Error Essence: [...]\\
\hspace*{4pt}Error Type: [...]\\
\hspace*{4pt}Error Mechanism: [...]\\
\hspace*{4pt}Error Impact: [...]\\
\hspace*{4pt}Improvement Suggestion: [...]
\end{promptbox}
\caption{Meta-prompt for \textbf{Step 1: Open-Ended Error Diagnosis}. It enables unbiased identification of root causes and failure mechanisms prior to targeted prompt optimization.}
\label{fig:prompt_diagnose_1}
\end{figure*}

\begin{figure*}[h]
\centering
\begin{promptbox}[title={\textsc{Meta-Prompt: Step 2 - Factor Selection}}]
\small\ttfamily
You are an expert in prompt optimization and error diagnosis. Your goal is to select the SINGLE most relevant factor to improve, based strictly on the given error analysis.

\vspace{2pt}
\colorbox{blue!8}{\textbf{Available Factors}}\\
\hspace*{4pt}(choose exactly ONE from this list)\\
\hspace*{4pt}\{factor\_list\}

\vspace{2pt}
\colorbox{blue!8}{\textbf{Error Analysis Summary}}\\
\hspace*{4pt}Primary Error Type: \{error\_type\}\\
\hspace*{4pt}Total Wrong Examples: \{num\_errors\}

\vspace{2pt}
\colorbox{blue!8}{\textbf{Representative Error Samples}}\\
\hspace*{4pt}\{error\_samples\}

\vspace{2pt}
\colorbox{blue!8}{\textbf{Selection Criteria}}\\
\hspace*{4pt}1. The factor whose improvement would most directly resolve the root cause.\\
\hspace*{4pt}2. The factor whose scope best aligns with the observed failure patterns.\\
\hspace*{4pt}3. The factor with the highest potential to prevent similar future errors.

\vspace{2pt}
\colorbox{yellow!25}{\textbf{Factor Selection History}}\\
\hspace*{4pt}Frequently selected: \{overexplored\_factors\}\\
\hspace*{4pt}Less explored: \{underexplored\_factors\}\\
\hspace*{4pt}\textit{Recommendation: If the less explored factors are also relevant to}\\
\hspace*{4pt}\textit{solving the current errors, consider giving them opportunities to}\\
\hspace*{4pt}\textit{ensure balanced exploration and avoid over-focusing on a single factor.}

\vspace{2pt}
\colorbox{red!15}{\textbf{Output Constraints}}\\
\hspace*{4pt}Output ONLY the factor name.\\
\hspace*{4pt}The output must exactly match one item in the factor list.\\
\hspace*{4pt}Do NOT include explanations or additional text.
\end{promptbox}
\caption{Meta-prompt for \textbf{Step 2: Factor Selection}. It maps diagnosed errors to specific factors, with history-aware recommendations to encourage balanced exploration when a single factor is repeatedly selected.}
\label{fig:prompt_select_2}
\end{figure*}

\subsection{Error Diagnosis and Factor Selection}
\label{app:metaprompts:diagnose}

aPSF uses a two-stage error analysis pipeline: (1) open-ended error diagnosis and (2) diagnosis-conditioned factor selection. See Figures~\ref{fig:prompt_diagnose_1} and \ref{fig:prompt_select_2} for the complete meta-prompts.

\subsection{Meta-Prompt Design Principles}

Our meta-prompts follow these design principles:

\paragraph{Clarity and specificity.} Each meta-prompt clearly defines the task, provides explicit formatting requirements, and includes concrete examples where helpful.

\paragraph{Constraint specification.} We specify both positive requirements (what to do) and negative constraints (what to avoid), reducing ambiguity in LLM responses.

\paragraph{Structured output.} All meta-prompts enforce structured output formats using delimiters (e.g., \texttt{===}, \texttt{---}), enabling reliable parsing.

\paragraph{Few-shot guidance.} Initial-prompt analysis includes a concrete example showing the input-output transformation, improving Architect LLM performance.

\paragraph{Factor-type awareness.} Edit meta-prompt provides type-specific guidance for different factor types (Instruction, Examples, Rationale, etc.), tailoring optimization strategies.

\section{Detailed BBH Results}
\label{app:bbh_detail}

Table~\ref{tab:complex_reasoning} highlights five sub-tasks where aPSF achieves exceptional performance (accuracy $\ge$ 86\%) and significant margins over the strongest baselines. Notably, on \textit{Multistep Arithmetic Two}, aPSF achieves \textbf{96.0\%} accuracy compared to 87.0\% for OPRO, demonstrating the power of separating calculation from formatting.

\begin{table*}[ht]
\centering
\small
\renewcommand{\arraystretch}{1.12}
\setlength{\tabcolsep}{5pt}
\begin{tabular}{lcccccc}
\toprule
\textbf{BBH Sub-task} & \textbf{aPSF} & \textbf{ProTeGi} & \textbf{OPRO} & \textbf{GrIPS} & \textbf{APE} & \makecell{\textbf{Zero-shot}\\\textbf{CoT}} \\
\midrule
boolean\_expressions & \textbf{97.0} & 95.5 & 93.5 & 94.5 & 94.5 & 93.0 \\
multistep\_arithmetic\_two & \textbf{96.0} & 86.0 & 87.0 & 86.0 & 84.0 & 81.0 \\
reasoning\_about\_colored\_objects & \textbf{92.0} & 76.0 & 77.0 & 71.0 & 66.0 & 72.0 \\
web\_of\_lies & \textbf{90.5} & 85.0 & 84.0 & 86.0 & 85.5 & 81.5 \\
tracking\_shuffled\_objects\_three & \textbf{86.0} & 85.0 & 77.0 & 78.0 & 84.0 & 70.0 \\
\midrule
\textit{Average (Top-5)} & \textbf{92.3} & 85.5 & 83.7 & 83.1 & 82.8 & 79.5 \\
\bottomrule
\end{tabular}
\caption{Top-performing BBH sub-tasks. aPSF demonstrates decisive advantages on arithmetic, boolean logic, and object tracking tasks, validating the efficacy of the \textsc{Format} and \textsc{Rationale} factors.}
\label{tab:complex_reasoning}
\end{table*}

\section{Performance Profile Analysis}
\label{app:performance_profile}

\begin{figure}[h]
\centering
\includegraphics[width=0.95\linewidth]{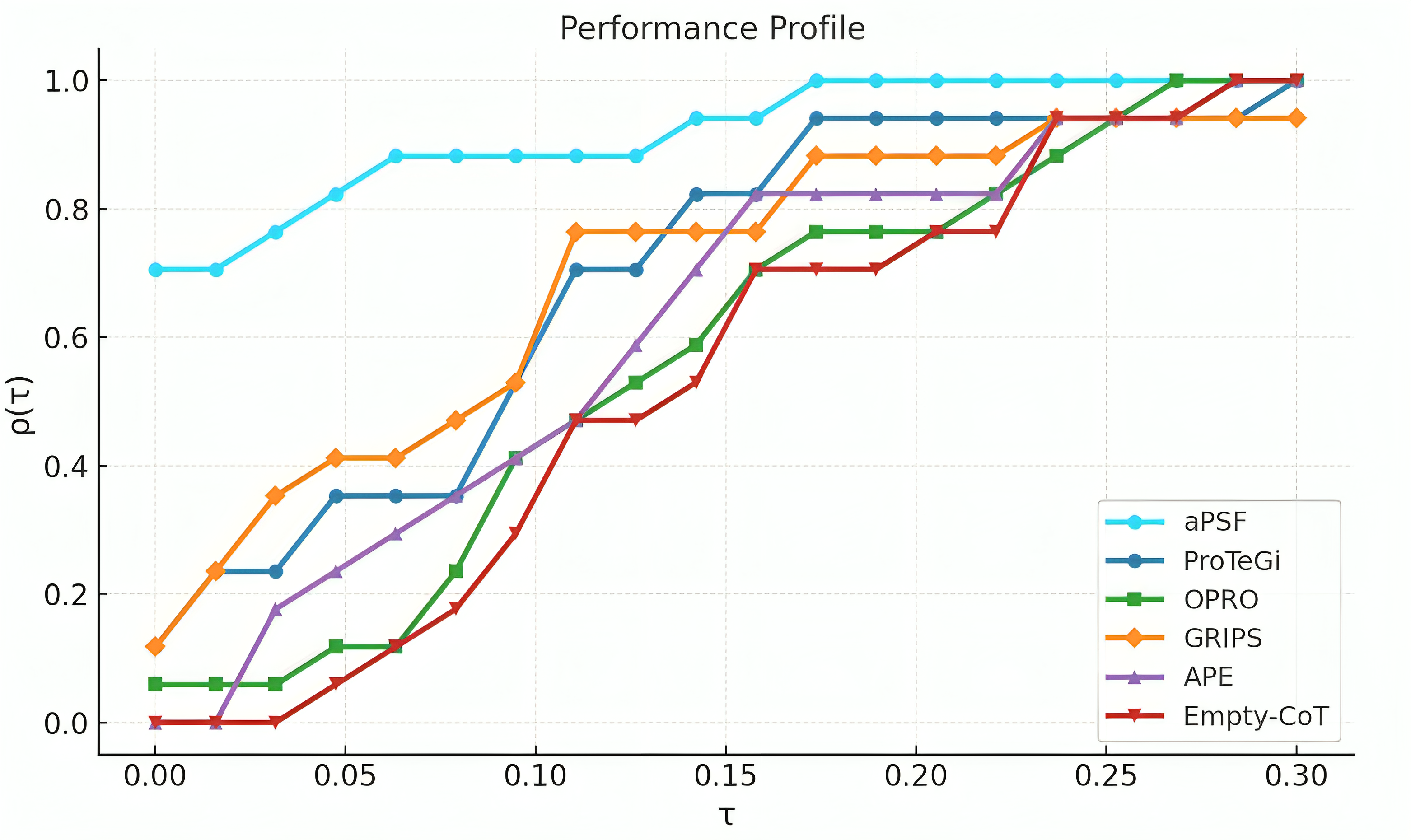}
\caption{
Performance profile on BBH sub-tasks.
aPSF (red) exhibits the steepest curve and highest plateau, indicating it most frequently achieves near-optimal performance across tasks.
At $\tau{=}0.1$ (within 10\% of best), aPSF succeeds on 76\% of tasks vs.\ 65\% for OPRO. Empty-CoT denotes the initial prompt baseline without prompt optimization.}

\label{fig:performance_profile}
\end{figure}

Figure~\ref{fig:performance_profile} shows performance profiles across the 17 BBH sub-tasks we evaluate. The performance profile plots the fraction of tasks where each method achieves accuracy within a factor $\tau$ of the best method. aPSF exhibits the steepest curve and highest plateau, indicating it most frequently achieves near-optimal performance. At $\tau{=}0.1$ (within 10\% of best), aPSF succeeds on 76\% of tasks compared to 65\% for OPRO, demonstrating superior robustness across diverse reasoning categories.

\section{MMLU Results}
\label{app:mmlu_results}

To evaluate aPSF on knowledge-intensive tasks, we test on MMLU~\citep{hendrycks2020measuring} (5-shot, 57 subjects). Table~\ref{tab:mmlu_top_grouped_no_subset} shows aPSF achieves the best overall average (76.29\%), with strongest improvement on Humanities (+6.18pp) and Other (+1.10pp), though it does not lead in Social Sciences or STEM categories.

\begin{table*}[t]
\centering
\small
\setlength{\tabcolsep}{5pt}
\renewcommand{\arraystretch}{1.12}

\textbf{\Large MMLU Results}\\[2pt]

\begin{tabular}{l rrrrrrr}
\toprule
\textbf{Category}
& \textbf{aPSF} & \textbf{OPRO} & \textbf{GrIPS}
& \textbf{ProTeGi} & \textbf{APE} & \makecell{\textbf{Zero-shot}\\\textbf{CoT}} & \textbf{$\Delta$ vs Best Baseline} \\
\midrule

Humanities
& \cellcolor{lightblue}\textbf{67.98} & \underline{61.80} & 60.91 & 60.84 & 59.31 & 55.25 & \textcolor{darkgreen}{\textbf{+6.18}} \\

Social Sciences
& 80.04 & \underline{80.70} & \cellcolor{lightblue}\textbf{81.19} & 80.34 & 79.94 & 70.83 & \textcolor{red}{\textbf{-1.15}} \\

STEM
& 76.75 & \cellcolor{lightblue}\textbf{77.80} & \underline{77.54} & 77.12 & 75.98 & 66.42 & \textcolor{red}{\textbf{-1.05}} \\

Other
& \cellcolor{lightblue}\textbf{80.37} & 78.60 & 78.35 & \underline{79.27} & 77.08 & 68.91 & \textcolor{darkgreen}{\textbf{+1.10}} \\
\midrule

\rowcolor{lightpurple}
Overall Avg.
& \textbf{76.29} & \underline{74.72} & 74.50 & 74.39 & 73.08 & 65.35 & \textcolor{darkgreen}{\textbf{+1.57}} \\
\bottomrule
\end{tabular}

\caption{MMLU accuracy (\%) across four disciplinary categories. aPSF achieves best overall average (76.29\%) with strongest gains on Humanities (+6.18pp) and Other (+1.10pp), though trailing in Social Sciences and STEM. $\Delta$ vs Best Baseline = aPSF's difference from the strongest non-aPSF baseline in each category (\textcolor{darkgreen}{green}=aPSF wins, \textcolor{red}{red}=aPSF loses). Light blue highlights indicate category winners; purple row emphasizes overall average.}
\label{tab:mmlu_top_grouped_no_subset}
\end{table*}

\section{Additional Benchmark Results: GPQA and Competition Math}
\label{app:gpqa_math_results}

We further evaluate aPSF on scientific reasoning (GPQA~\citep{rein2024gpqa}) and advanced mathematical reasoning (Competition Mathematics~\citep{hendrycks2021measuring}, 500 test samples). Table~\ref{tab:gpqa_math} shows aPSF achieves consistent improvements on these challenging benchmarks.

\begin{table}[h]
\centering
\footnotesize
\setlength{\tabcolsep}{2.5pt}
\begin{tabular}{l ccccc}
\toprule
& \textbf{aPSF} & \textbf{OPRO} & \textbf{APE} & \textbf{ProTeGi} & \textbf{GrIPS} \\
\midrule
Comp.\ Math & \cellcolor{lightblue}\textbf{56.0} & 51.6 & 54.0 & 54.0 & \underline{55.0} \\
GPQA-Chem & \cellcolor{lightblue}\textbf{30.2} & 23.8 & 20.6 & 19.1 & \underline{28.6} \\
GPQA-Phys & \cellcolor{lightblue}\textbf{45.9} & \underline{44.3} & 41.0 & 39.3 & 39.3 \\
\bottomrule
\end{tabular}
\caption{Test accuracy (\%) on Competition Math and GPQA. Comp.\ Math = Competition Mathematics; Chem = Chemistry; Phys = Physics. aPSF achieves best on all benchmarks.}
\label{tab:gpqa_math}
\end{table}

\section{Hyperparameter Sensitivity Analysis}
\label{app:hyperparam}

We evaluate aPSF's sensitivity to the candidate count $N$ per factor update on the challenging GSM-Hard benchmark. As shown in Table~\ref{tab:n_ablation}, performance does not increase monotonically with $N$.

\begin{table}[h]
\centering
\small
\renewcommand{\arraystretch}{1.1} 
\setlength{\tabcolsep}{4pt}       
\begin{tabular}{cc l}             
\toprule
\textbf{Count ($N$)} & \textbf{GSM-Hard (\%)} & \textbf{Trend} \\
\midrule
2 & 52.78 & Under-exploration \\
\textbf{4} & \textbf{54.77} & \textbf{Optimal Balance} \\
6 & 53.88 & Diminishing Returns \\
8 & 50.80 & Overfitting / Noise \\
\bottomrule
\end{tabular}
\caption{Sensitivity to candidate count $N$ on GSM-Hard. Performance peaks at $N=4$, while larger $N$ degrades performance.}
\label{tab:n_ablation}
\end{table}

\textbf{Analysis:}
\begin{itemize}[leftmargin=*, itemsep=2pt]
\item \textbf{Under-exploration ($N=2$)}: Limited candidates fail to uncover effective local edits, resulting in suboptimal performance (52.78\%).
\item \textbf{Optimal Balance ($N=4$)}: This setting achieves the peak accuracy (54.77\%), effectively balancing exploration width with selection stability.
\item \textbf{Overfitting Risk ($N \ge 6$)}: Contrary to the intuition that ``more is better,'' increasing $N$ to 8 causes a significant drop to 50.80\%. We hypothesize that a larger candidate pool increases the likelihood of finding ``false positives'' prompts that exploit artifacts in the small validation slice (50 examples) but fail to generalize to the test set. Thus, we adopt $N=4$ as the robust default.
\end{itemize}

\section{From-Scratch Capability (No Initial Prompt)}
\label{app:initial_prompt}

Our main results use \emph{initial-prompt mode} (Section~\ref{sec:setup}). Here we evaluate aPSF in \emph{from-scratch mode}, where the Architect discovers a factorized prompt without any user-provided seed.

Table~\ref{tab:mode_comparison} compares the two aPSF modes directly.

\begin{table}[h]
\centering
\small
\renewcommand{\arraystretch}{1.15}
\setlength{\tabcolsep}{4pt}
\resizebox{0.9\linewidth}{!}{
    \begin{tabular}{lccc}
    \toprule
    \textbf{Benchmark} & \textbf{aPSF (Initial)} & \textbf{aPSF (Scratch)} & \textbf{$\Delta$} \\
    \midrule
    GSM-Hard & \textbf{54.77} & 54.67 & $-0.10$ \\
    AQUA-RAT & 82.50 & \textbf{83.00} & $+0.50$ \\
    MultiArith & \textbf{99.30} & \textbf{99.30} & $\pm 0.00$ \\
    \bottomrule
    \end{tabular}
}
\caption{Comparison of aPSF in \textbf{Initial-Prompt} mode (``Let's think step by step'') vs.\ \textbf{From-Scratch} mode. Performance is comparable, confirming that aPSF's gains stem from the factorization mechanism rather than the initial prompt.}
\label{tab:mode_comparison}
\end{table}

\paragraph{Key findings.}
\begin{itemize}[leftmargin=*, itemsep=2pt]
    \item \textbf{Comparable performance:} The two modes achieve nearly identical results across all three benchmarks, with differences within 0.5pp. This confirms that aPSF's structure discovery is robust regardless of initialization.
    \item \textbf{From-scratch can match or exceed:} On \textbf{AQUA-RAT}, from-scratch mode slightly outperforms initial-prompt mode (+0.50pp), suggesting that autonomously discovered structures can be as effective as human-guided ones.
    \item \textbf{Gains not solely from seed:} These results confirm that aPSF's improvements stem from the factorization and optimization mechanism itself, not merely from a privileged starting point.
\end{itemize}

\section{Complete Factor Selection Statistics}
\label{app:factor_stats}

Table~\ref{tab:factor_complete} presents comprehensive factor selection statistics across all evaluated tasks. The test accuracy figures align with the main results reported in Section~\ref{sec:results}.

\begin{table*}[ht]
\centering
\small
\renewcommand{\arraystretch}{1.12}
\setlength{\tabcolsep}{4pt}
\begin{tabular}{lccccl}
\toprule
\textbf{Task} & \textbf{Test Acc (\%)} & \textbf{Val Gain} & \textbf{Gen Gap} & \textbf{Dominant Factor} & \textbf{Selection Rate} \\
\midrule
\multicolumn{6}{l}{\textit{Mathematical Reasoning:}} \\
GSM8K & \textbf{90.03} & +0.0\% & 0.020 & Result Aggregation & 50\% \\
AQUA-RAT & \textbf{82.50} & +10.3\% & 0.035 & Calculation Execution & 50\% \\
GSM-Hard & \textbf{54.77} & +10.3\% & 0.102 & Problem Understanding & 67\% \\
MultiArith & \textbf{99.30} & +0.0\% & -0.033 & Problem Analysis & 100\% \\
\midrule
\multicolumn{6}{l}{\textit{Logical Reasoning (BBH):}} \\
Date Understanding & \textbf{80.00} & +12.2\% & 0.167 & Time Difference Calc & 100\% \\
Logical Deduction & \textbf{75.00} & +8.6\% & 0.013 & Establishing Order & 80\% \\
\midrule
\multicolumn{6}{l}{\textit{Scientific Reasoning (GPQA):}} \\
Chemistry & 30.16 & +100\% & 0.098 & Reagent Evaluation & 50\% \\
Physics & 45.90 & +200\% & 0.166 & Principle Application & 70\% \\
Biology & 50.00 & +0.0\% & 0.250 & Sci. Principle App. & 50\% \\
\midrule
\multicolumn{6}{l}{\textit{High-Complexity Tasks:}} \\
AIME-2025 & 8.33 & +100\% & 0.250 & Parse Problem & 50\% \\
\bottomrule
\end{tabular}
\caption{Complete factor selection statistics across all tasks. \textbf{Test Acc} matches the main results using the default configuration (Qwen2.5-7B-Instruct Worker + Qwen3-8B Architect). Val Gain = validation improvement from initial to best; Gen Gap = difference between validation and test accuracy.}
\label{tab:factor_complete}
\end{table*}

\paragraph{Key observations from factor selection patterns.}
\begin{itemize}[leftmargin=*, itemsep=2pt]
\item \textbf{Task-specific adaptation}: Error-guided selection identifies domain-appropriate factors---mathematical tasks often prioritize Calculation or Problem Understanding, logic tasks focus on Step-by-Step reasoning, and scientific tasks emphasize Principle Application.
\item \textbf{Selection Concentration}: Tasks with 100\% selection on a single factor indicate two distinct states: \textbf{saturation} (e.g., MultiArith, +0.0\% gain) where the initial prompt is already optimal, or \textbf{critical bottlenecking} (e.g., BBH-Date, +12.2\% gain) where one specific skill dictates performance.

\begin{figure}[h]
\centering
\includegraphics[width=0.9\linewidth]{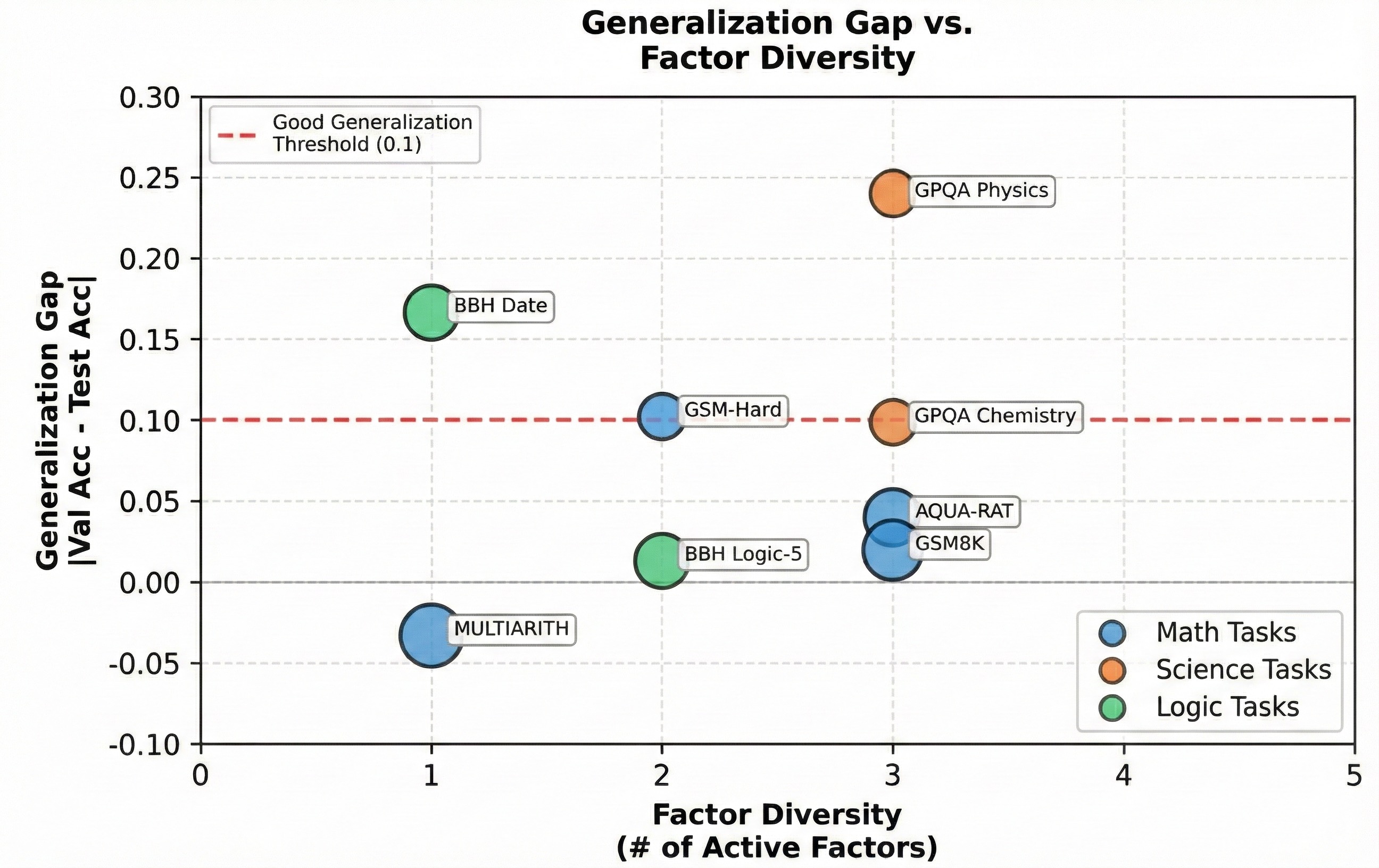} 
\caption{
    \textbf{Generalization Gap vs. Factor Diversity.}
    Visualizing the relationship between factor count and generalization gap (data from Table~\ref{tab:factor_complete}).
    Robust tasks (e.g., GSM8K, AQUA) maintain low gaps ($<0.05$) with balanced factors. High gaps in complex tasks (e.g., GPQA) reflect intrinsic difficulty rather than optimization overfitting.
}
\label{fig:app_gen_gap}
\end{figure}

\item \textbf{Generalization quality}: \textbf{As visualized in Figure~\ref{fig:app_gen_gap}}, tasks with balanced factor distributions (20--50\% each) exhibit lower generalization gaps (GSM8K: 0.020, AQUA: 0.035), whereas concentrated or high-complexity tasks correlate with higher gaps (AIME: 0.250).

\item \textbf{Validation-test alignment}: High validation gains (+100--200\%) on challenging tasks (GPQA-Physics, AIME) indicate effective optimization, confirming that the framework can exploit headroom even when absolute accuracy remains limited by task difficulty.
\end{itemize}

\section{Detailed Factor Names by Task}
\label{app:factor_names}

Table~\ref{tab:factor_names_detailed} presents the original, task-specific factor names discovered by aPSF's Architect LLM.

\begin{table*}[ht]
\centering
\small
\renewcommand{\arraystretch}{1.2} 
\setlength{\tabcolsep}{6pt}

\begin{tabularx}{\textwidth}{l X l}
\toprule
\textbf{Task} & \textbf{Discovered Factor Names (Original)} & \textbf{Top Factor (Rate)} \\
\midrule
\multicolumn{3}{l}{\textit{Mathematical Reasoning:}} \\
GSM8K & StepByStepDecomposition, ComponentAnalysis, CalculationExecution, ResultAggregation & ResultAggregation (50\%) \\
AQUA-RAT & StepByStepReasoning, ProblemAnalysis, StepBreakdown, CalculationExecution, Verification, AnswerSelection & CalculationExecution (50\%) \\
GSM-Hard & ProblemUnderstanding, StepByStepBreakdown, MathematicalOperations, FinalCalculation, Verification & ProblemUnderstanding (67\%) \\
MultiArith & StepByStepApproach, ProblemAnalysis, SequentialExecution, ResultValidation & ProblemAnalysis (100\%) \\
\midrule
\multicolumn{3}{l}{\textit{Logical Reasoning (BBH):}} \\
Date Understanding & StepByStepApproach, DateParsing, TimeDifferenceCalculation, CalendarRuleApplication, OptionVerification & TimeDifferenceCalculation (100\%) \\
Logical Deduction & ParsingStatements, EstablishingOrder, EvaluatingOptions & EstablishingOrder (80\%) \\
\midrule
\multicolumn{3}{l}{\textit{Scientific Reasoning (GPQA):}} \\
Chemistry & ReactionAnalysis, ReagentEvaluation, ProductStability, StereochemistryConsideration, OptionElimination & ReagentEvaluation (50\%) \\
Physics & ContextAnalysis, ParameterExtraction, PrincipleApplication, CalculationExecution, ValidationCheck & PrincipleApplication (70\%) \\
Biology & ContextualInterpretation, ScientificPrincipleApplication, OptionEvaluation, Elimination\&Justification & Sci.\ Principle App.\ (50\%) \\
\midrule
\multicolumn{3}{l}{\textit{High-Complexity Tasks:}} \\
AIME-2025 & ParseProblem, IdentifyConcepts, BreakSubproblems, StepwiseReasoning, ValidateSteps, SynthesizeResults & ParseProblem (50\%) \\
\bottomrule
\end{tabularx}
\caption{Original factor names autonomously discovered by aPSF. The terminology adapts to the domain (e.g., \textit{Stereochemistry} in Chemistry, \textit{CalculationExecution} in AQUA). Top Factor indicates the component most frequently selected by the error-guided scheduler.}
\label{tab:factor_names_detailed}
\end{table*}

\paragraph{Semantic grouping for cross-task analysis.}
For the heatmap visualization in Figure~\ref{fig:factor_patterns} (main text), we semantically group task-specific factors into six general categories to enable cross-task pattern comparison: (\textit{i})~\textbf{Problem Understanding} (e.g., ProblemAnalysis, ComponentAnalysis, ParsingStatements), (\textit{ii})~\textbf{Step-by-Step Breakdown} (e.g., EstablishingOrder, StepByStepBreakdown), (\textit{iii})~\textbf{Mathematical Operations} (e.g., MathematicalOperations, FormulaUsage), (\textit{iv})~\textbf{Calculation Execution} (e.g., CalculationExecution, TimeDifferenceCalculation), (\textit{v})~\textbf{Verification \& Validation} (e.g., Verification, ResultValidation), and (\textit{vi})~\textbf{Result Aggregation} (e.g., ResultAggregation, AnswerSelection, LogicalDeduction). This grouping preserves semantic intent while enabling macro-level pattern analysis across diverse task domains.

\section{Factor Discovery Examples}
\label{app:examples}

This section provides concrete examples of discovered factor structures and illustrates the prompt transformation process.

\subsubsection{Initial-Prompt Mode Extension Example}

Starting from the baseline ``Let's think step by step'' on GSM8K, the Architect LLM analyzes the reasoning style and generates the following extension:

\begin{quote}
\small
\textbf{Extended Instruction:} ``Let's think step by step, carefully analyzing the problem's components and their relationships, performing each calculation with clear intermediate steps, and logically combining all results to arrive at the final answer.''
\end{quote}

The system then decomposes this into independently optimizable factors:
\begin{itemize}[leftmargin=*, itemsep=1pt]
\item \textbf{$\mathcal{F}_1$-StepByStepDecomposition}: ``Let's think step by step''
\item \textbf{$\mathcal{F}_2$-ComponentAnalysis}: ``carefully analyzing the problem's components and their relationships''
\item \textbf{$\mathcal{F}_3$-CalculationExecution}: ``performing each calculation with clear intermediate steps''
\item \textbf{$\mathcal{F}_4$-ResultAggregation}: ``logically combining all results to arrive at the final answer''
\end{itemize}

This demonstrates how aPSF preserves the core reasoning style while injecting task-specific procedural guidance, enabling subsequent factor-wise optimization.

\subsubsection{BBH--Logical Deduction: A Complete Optimization Trace}

We present a complete example on BBH--Logical Deduction (Five Objects), a task requiring constraint parsing, order establishment, and option verification. Starting from the same initial prompt baseline (Zero-shot):

\begin{quote}
\small
\textbf{Initial Prompt:} ``Let's think step by step.''
\end{quote}

The Architect LLM analyzes the task requirements (parsing comparative statements like ``A is before B'', building a global ordering chain) and generates:

\begin{quote}
\small
\textbf{Extended Instruction:} ``Carefully read the given statements and identify all comparative relationships between the items. Use these relationships to create a logical order of the items from highest to lowest. Then, evaluate each option against this order to determine which one is correct. When establishing the order, build the sequence incrementally by validating each comparative relationship against the current order and explicitly confirming chains of relationships (e.g., A < B < C) step by step to prevent logical gaps.''
\end{quote}

The system decomposes this into three semantic factors:
\begin{itemize}[leftmargin=*, itemsep=1pt]
\item \textbf{$\mathcal{F}_1$-ParsingStatements}: ``Carefully read the given statements and identify all comparative relationships between the items.''
\item \textbf{$\mathcal{F}_2$-EstablishingOrder}: ``Use these relationships to create a logical order... build the sequence incrementally by validating each comparative relationship...''
\item \textbf{$\mathcal{F}_3$-EvaluatingOptions}: ``Evaluate each option against this order to determine which one is correct.''
\end{itemize}

\paragraph{Error-guided optimization results.}
During optimization, error analysis reveals that 80\% of failures stem from incorrect order establishment (e.g., missing transitive relationships, premature chain termination). The error-guided selector consequently prioritizes $\mathcal{F}_2$-EstablishingOrder across 8 of 10 optimization rounds. Key metrics:

\begin{center}
\small
\begin{tabular}{lc}
\toprule
\textbf{Metric} & \textbf{Value} \\
\midrule
Initial validation accuracy & 70.0\% \\
Best validation accuracy & 76.0\% \\
Final test accuracy & 74.67\% \\
Validation improvement & +8.6\% \\
Generalization gap & 0.013 (excellent) \\
Dominant factor & EstablishingOrder (80\%) \\
\bottomrule
\end{tabular}
\end{center}

This example illustrates how aPSF automatically identifies task-specific bottlenecks and focuses optimization effort accordingly.

\subsubsection{Discovered Factor Structures by Task Type}

Table~\ref{tab:factor_examples} presents concrete factor structures autonomously discovered by aPSF for diverse task types, illustrating adaptive factorization based on task complexity. Note that factor names align with the terminology identified in Table~\ref{tab:factor_names_detailed}.

\begin{table*}[ht]
\centering
\small
\renewcommand{\arraystretch}{1.3}
\setlength{\tabcolsep}{6pt}

\begin{tabularx}{\textwidth}{l X c}
\toprule
\textbf{Task Type} & \textbf{Discovered Factors} & \textbf{Order} \\
\midrule
Math (GSM8K)
& 1. StepByStepDecomposition: Decompose problem step by step \newline
  2. ComponentAnalysis: Identify key quantities \newline
  3. CalculationExecution: Perform arithmetic operations \newline
  4. ResultAggregation: Format final output
& $1 \to 2 \to 3 \to 4$ \\
\midrule
Logic (BBH)
& 1. ParsingStatements: Extract logical premises \newline
  2. EstablishingOrder: Deduce chronological/logical sequence \newline
  3. EvaluatingOptions: Verify consistency against premises \newline
  4. OptionVerification: Double-check constraints
& $1 \to 2 \to 3 \to 4$ \\
\midrule
QA (AQUA)
& 1. ProblemAnalysis: Identify question type \newline
  2. CalculationExecution: Perform systematic calculations \newline
  3. Verification: Self-validate plausibility \newline
  4. AnswerSelection: Select correct option
& $1 \to 2 \to 3 \to 4$ \\
\bottomrule
\end{tabularx}
\caption{Factor structures discovered by aPSF. The factor names (e.g., \textit{ComponentAnalysis}, \textit{CalculationExecution}) correspond to the domain-specific terminology cataloged in Appendix~\ref{app:factor_names}.}
\label{tab:factor_examples}
\end{table*}

\section{Theoretical Analysis}
\label{app:theory}

We provide a formal intuition for why aPSF's error-guided factor selection outperforms fixed schedules, focusing on regret minimization under non-stationary rewards.

\subsection{Problem Formulation}
We view prompt optimization as a sequential decision process. At step $t$, the Architect chooses a factor $k \in \{1,\dots,K\}$ to optimize. Let $\mu_k^{(t)}$ be the potential improvement gain of factor $k$ at time $t$. The cumulative regret after $T_{\max}$ steps is:
\begin{equation}
R(T_{\max}) = \sum_{t=1}^{T_{\max}} \max_k(\mu_k^{(t)}) - \sum_{t=1}^{T_{\max}} \mu_{k_t}^{(t)},
\end{equation}
where $\max_k(\mu_k^{(t)})$ is the gain of the optimal action (Oracle).

\subsection{Why Error-Guided Factor Selection Wins}

\paragraph{Fixed policies (Round-robin).}
A round-robin policy selects $k$ uniformly, ignoring the current state. If factor potentials are heterogeneous (e.g., \textsc{Format} is broken while \textsc{Instruction} is perfect), fixed policies waste $1 - 1/K$ of the budget on saturated factors, leading to linear regret.

\paragraph{Error-Guided Factor Selection as Contextual Bandits.}
Our Error-Guided factor selection policy $\pi$ observes the current \textbf{error distribution} $\mathcal{E}^{(t)}$ (Context) to select action $k$.
Under the assumption that the Architect's diagnosis accuracy is $\rho > 1/K$, the policy prioritizes high-potential factors (e.g., fixing \textsc{Format} when format errors are high).
This effectively creates a shortcut to convergence:
\begin{equation}
\E_{\pi}[\text{gain}] \approx \rho \cdot \mu_{\text{bottleneck}}^{(t)} + (1-\rho) \cdot \bar{\mu}^{(t)}.
\end{equation}
By focusing on the bottleneck, $R(T_{\max})$ is minimized significantly faster than round-robin, consistent with the rapid convergence observed in the trajectory analysis (\textbf{Figure~\ref{fig:bbh_date_stepwise} in Appendix~\ref{app:trajectory}}).

\subsection{Independence Assumption and Complexity}

aPSF optimizes factors iteratively (Coordinate Descent), assuming partial independence: $\hat{S}(A \cup B) \approx \hat{S}(A) + \hat{S}(B)$.
While subtle factor interactions naturally exist (e.g., between reasoning steps and formatting constraints), modeling the full joint distribution would result in an exponential search space $O(C^K)$ (where $C$ is the candidate count).
By adopting the independence assumption, aPSF reduces the complexity to linear $O(C \cdot K \cdot T_{\max})$. This trade-off makes optimization tractable for LLMs, as demonstrated by the strong empirical results across diverse benchmarks.

\subsection{Convergence Properties Under Interventional Updates}

\textbf{Note}: This section provides informal convergence arguments rather than rigorous proofs.

Let $B^{(t)}$ denote the factor contents at iteration $t$, and $\hat{S}^{(t)} = \hat{S}(\mathrm{Assemble}(B^{(t)}), \mathcal{D}_{\mathrm{val}})$ be the validation accuracy. aPSF's update rule is:
\begin{equation}
f_k^{(t+1)} = \begin{cases}
\hat{c}_k & \text{if } \Delta \hat{S}(\hat{c}_k) \ge \delta \\
f_k^{(t)} & \text{otherwise}
\end{cases},
\end{equation}
where $\hat{c}_k$ is the best candidate proposed by the Architect $M_A$, and $\delta > 0$ is the acceptance threshold.

\paragraph{Monotonicity.} By construction, $\hat{S}^{(t+1)} \ge \hat{S}^{(t)}$ (non-decreasing accuracy). Unlike non-contrastive methods that evaluate the whole prompt blindly, aPSF accepts updates only if the isolated marginal contribution $\Delta \hat{S} \ge \delta$. This strictly prevents regressions caused by noisy edits.

\paragraph{Convergence Intuition.} Assuming a bounded factor space and a capable Architect $M_A$ that generates diverse candidates, the system performs a hill-climbing search (Coordinate Descent). Since the objective function is bounded ($\hat{S} \in [0,1]$) and the sequence is non-decreasing, the optimization is expected to converge to a local maximum where no single-factor edit yields improvement $\ge \delta$.

\subsection{Information-Theoretic Justification}

Let $\mathcal{E}^{(t)}$ denote the distribution of validation errors at step $t$ (e.g., 40\% Arithmetic, 10\% Format).

\paragraph{Information-Theoretic Intuition.} Ideally, we want to select the factor $k$ whose edits are most informative about (and most likely to reduce) the current error modes. Formally, this corresponds to maximizing mutual information $I(\mathcal{F}_k; \mathcal{E}^{(t)})$ between factor $k$ and the error distribution. We do not estimate mutual information explicitly; instead, our \textbf{Error-Guided Factor Selection} approximates this criterion by using the Architect LLM to diagnose which factor is causally linked to the dominant error types.

\paragraph{Connection to Contextual Bandits.}
Standard prompt optimizers (like OPRO) often behave like Multi-Armed Bandits (MAB), exploring factors to maximize reward. However, standard MAB assumes stationary rewards. In prompt engineering, rewards are non-stationary (fixing the Instruction changes the utility of fixing Examples).
aPSF treats this as a \textbf{Contextual Bandit} problem:
\begin{itemize}
    \item \textbf{Context:} Current error distribution $\mathcal{E}^{(t)}$.
    \item \textbf{Action:} Select factor $k$.
    \item \textbf{Policy:} The Architect maps context (e.g., "Format Error") to action (e.g., "Fix Factor 4").
\end{itemize}
This contextual awareness allows aPSF to "exploit" known bottlenecks immediately, rather than wasting steps "exploring" saturated factors. This explains the rapid convergence observed in the trajectory analysis (\textbf{Figure~\ref{fig:bbh_date_stepwise} in Appendix~\ref{app:trajectory}}).

\section{Model Configurations}
\label{app:model_config}

\subsection{LLM Specifications}

To ensure reproducibility and accessibility, we primarily utilize high-performance open-weight models. We distinguish between two roles:

\paragraph{Worker Models ($M_W$).}
  The worker executes the task based on the assembled prompt. We evaluate two families:
  \begin{itemize}[leftmargin=*, itemsep=2pt]
  \item \textbf{Qwen2.5-7B-Instruct}: This is a 7B parameter instruction-tuned model, context window up to 32K tokens, quantization: FP16.
  \item \textbf{Llama-3.1-8B-Instruct}: This is an 8B parameter instruction-tuned model, context window up to 128K tokens, quantization: FP16.
  \end{itemize}

\paragraph{Architect LLM ($M_A$).}
The architect handles structure discovery, factor editing, and error diagnosis. We employ stronger models for these meta-reasoning tasks:
\begin{itemize}[leftmargin=*, itemsep=2pt]
\item \textbf{Qwen3-8B}: Used as the default Architect for most experiments due to its strong instruction-following capability.
\item \textbf{gpt-oss-120b}: A large-scale open-source model used in ablation studies to verify scalability. Accessed via API with temperature=0.7.
\end{itemize}

\subsection{Decoding Hyperparameters}

All inference uses the settings detailed in \textbf{Table~\ref{tab:decoding_params}}, unless explicitly varied.

\begin{table}[h]
\centering
\small
\begin{tabular}{lc}
\toprule
\textbf{Parameter} & \textbf{Value} \\
\midrule
Temperature & 0.0 (deterministic) \\
Top-p (nucleus sampling) & 1.0 (disabled) \\
Top-k & Not applied \\
Max output tokens & 8192 \\
Repetition penalty & 1.0 (disabled) \\
Stop sequences & \texttt{["\textbackslash n\textbackslash n", "---"]} \\
\bottomrule
\end{tabular}
\caption{Decoding hyperparameters for Worker model inference ($M_W$). These settings apply to both prompt selection on $\mathcal{D}_{\mathrm{val}}$ and final test evaluation.}
\label{tab:decoding_params}
\end{table}

For the Architect LLM ($M_A$) generating meta-level content (e.g., structure proposals, factor edits), we use \textbf{temperature=0.7} to encourage diversity while maintaining coherence.

\paragraph{Summary of Temperature Settings.}
\textbf{Worker} ($M_W$): temperature=0.0 (deterministic) for all inference, including both prompt selection and final test evaluation.
\textbf{Architect} ($M_A$): temperature=0.7 to encourage diverse structure proposals.

\section{Evaluation Protocol}
\label{app:eval_protocol}

\subsection{Data Splitting}

For each benchmark, we utilize a small, fixed \textbf{validation slice} ($|\mathcal{D}_{\mathrm{val}}| = 50$ examples) for prompt optimization and a held-out test set for final evaluation, as detailed in Table~\ref{tab:data_split}:

\begin{table}[h]
\centering
\small
\begin{tabular}{lcc}
\toprule
\textbf{Dataset} & \textbf{Validation Slice} & \textbf{Test Size} \\
\midrule
GSM8K & 50 & 1,319 \\
AQUA-RAT & 50 & 254 \\
MultiArith & 50 & 180 \\
GSM-Hard & 50 & 1006 \\
BBH (per task) & 50 & Variable \\
\bottomrule
\end{tabular}
\caption{Dataset splits. We use a fixed validation slice of 50 examples per task for prompt optimization. Test sets follow official benchmarks where available.}

\label{tab:data_split}
\end{table}

\subsection{Scoring Functions}

\paragraph{Math benchmarks (GSM8K, AQUA, MultiArith, GSM-Hard).}
We use \textbf{exact-match accuracy}: extract the final numerical answer using regex patterns (e.g., ``The answer is \textbackslash d+''), normalize to remove commas/units, and compare string equality with ground truth.

\paragraph{BBH and complex reasoning benchmarks.}
For tasks with diverse answer formats (e.g., Yes/No, multiple choice A--E, or free-form text), we use \textbf{LLM-based answer extraction}: a lightweight LLM extracts the final answer from model outputs, which is then compared against ground truth for exact-match accuracy. This approach handles varied response formats more robustly than regex patterns.

\section{Computational Resources}
\label{app:compute}

All experiments run on the following infrastructure:

\begin{itemize}[leftmargin=*, itemsep=2pt]
\item \textbf{GPUs}: 2$\times$ NVIDIA A100 80GB GPUs for worker inference
\item \textbf{CPU}: 64-core AMD EPYC 7763 @ 2.45GHz
\item \textbf{RAM}: 512GB DDR4
\item \textbf{OS}: Ubuntu 20.04 LTS, CUDA 11.8, PyTorch 2.0.1
\end{itemize}

\section{Datasets and Licenses}
\label{app:data}

We summarize the benchmarks, task types, split sources, and license details in \textbf{Table~\ref{tab:datasets}}. and include the following information for reproducibility:
\begin{itemize}[leftmargin=*, itemsep=2pt]
\item \textbf{Download URLs}: Provided in code release README.
\item \textbf{Preprocessing}: Scripts for answer normalization and format extraction are included in the supplementary material.
\item \textbf{Version Control}: Pinned commit hashes are used to ensure reproducibility.
\end{itemize}

\begin{table*}[t]
\centering
\small
\renewcommand{\arraystretch}{1.2}
\setlength{\tabcolsep}{12pt}

\begin{tabular}{lll}
\toprule
\textbf{Dataset} & \textbf{Task Type} & \textbf{Reference} \\
\midrule
GSM8K & Math word problems & \citet{cobbe2021training} \\
AQUA-RAT & Math multiple-choice questions & \citet{ling2017program} \\
MultiArith & Multi-step arithmetic reasoning & \citet{roy2015solving} \\
GSM-Hard & Hard math with large numbers & \citet{gao2022pal} \\
MATH & Competition mathematics & \citet{hendrycks2021measuring} \\
BBH & Diverse reasoning (17 tasks) & \citet{suzgun2023challenging} \\
GPQA & Graduate-level science QA & \citet{rein2024gpqa} \\
MMLU & General knowledge (57 subjects) & \citet{hendrycks2020measuring} \\
\bottomrule
\end{tabular}

\caption{Summary of evaluation benchmarks. All datasets are publicly available.}
\label{tab:datasets}
\end{table*}

\section{Implementation Details}
\label{app:impl}

\subsection{Prompt Assembly}

Factors are concatenated in the specified order via the $\mathrm{Assemble}$ operation (Eq.~\ref{eq:assemble}), separated by newlines for readability. Total prompt length is implicitly bounded by the worker LLM's context window (up to 32K tokens for Qwen2.5 per official configuration); in practice, generated prompts rarely approach this limit.

\subsection{Validation Slice Design}

We use a fixed per-task validation slice for prompt optimization (Appendix~\ref{app:eval_protocol}).

\subsection{Stopping Criteria}

aPSF terminates when: (\textit{i})~$T_{\max}$ iterations are reached, or (\textit{ii})~validation performance has not improved for 3 consecutive iterations (early stopping).

\subsection{Prompt Length Sensitivity Analysis}

Prior work has shown that smaller language models are sensitive to prompt length~\citep{wei2022chain}. To validate this, we conducted ablation studies on prompt length using \textbf{Llama-3.1-8B-Instruct} (to test generalization beyond our default Qwen model). Table~\ref{tab:prompt_length} shows a clear inverse relationship between prompt length and accuracy on GSM8K.

\begin{table}[h]
\centering
\small
\begin{tabular}{cc}
\toprule
\textbf{Prompt Length (tokens)} & \textbf{Accuracy (\%)} \\
\midrule
362  & \textbf{89.62} \\
500  & 83.50 \\
653  & 80.55 \\
1260 & 76.47 \\
\bottomrule
\end{tabular}
\caption{Impact of prompt length on Llama-3.1-8B-Instruct performance (GSM8K). Accuracy degrades significantly as prompts exceed 500 tokens.}
\label{tab:prompt_length}
\end{table}

These findings suggest that prompt conciseness is particularly important for smaller models. In our main experiments, we do not impose explicit length constraints in the meta-prompts, allowing the Architect LLM to generate prompts of varying lengths; the factor-wise optimization naturally tends to retain concise, effective phrasings through performance-based selection.

\subsection{Extended Efficiency Analysis}
\label{app:efficiency_extended}

We extend the efficiency analysis from MultiArith (Section~\ref{sec:cost_analysis}) to four datasets. Table~\ref{tab:token_best_step} reports the token consumption at the best validation step and the step number itself, under the default setup (Worker: Qwen2.5-7B-Instruct; Architect: Qwen3-8B).

\begin{table}[h]
\centering
\small
\renewcommand{\arraystretch}{1.12}
\setlength{\tabcolsep}{3pt}
\begin{tabular}{lcccc}
\toprule
\multirow{2}{*}{\textbf{Method}} & \multicolumn{4}{c}{\textbf{Tokens at Best Step (K)}} \\
\cmidrule(lr){2-5}
& MultiArith & GSM8K & AQUA & GSM-Hard \\
\midrule
ProTeGi & 1455 & 2143 & 1896 & 2850 \\
GrIPS & 1640 & 1932 & 1968 & 2378 \\
CriSPO & 1050 & 1612 & 1644 & 2268 \\
APE & 824 & 1206 & 948 & 1445 \\
ZERA & 460 & 910 & 529 & 1148 \\
OPRO & 377 & 850 & 593 & 1110 \\
\rowcolor{lightblue}
\textbf{aPSF} & \textbf{206} & \textbf{491} & \textbf{336} & \textbf{648} \\
\bottomrule
\end{tabular}

\vspace{6pt}

\renewcommand{\arraystretch}{1.12}
\setlength{\tabcolsep}{3pt}
\begin{tabular}{lcccc}
\toprule
\multirow{2}{*}{\textbf{Method}} & \multicolumn{4}{c}{\textbf{Best Step ($\downarrow$)}} \\
\cmidrule(lr){2-5}
& MultiArith & GSM8K & AQUA & GSM-Hard \\
\midrule
GrIPS & 10 & 9 & 10 & 10 \\
ProTeGi & 7 & 8 & 8 & 10 \\
CriSPO & 5 & 6 & 7 & 8 \\
APE & 6 & 7 & 6 & 8 \\
ZERA & 3 & 5 & 3 & 6 \\
OPRO & 2 & 4 & 3 & 5 \\
\rowcolor{lightblue}
\textbf{aPSF} & \textbf{1} & \textbf{3} & \textbf{2} & \textbf{4} \\
\bottomrule
\end{tabular}
\caption{Extended efficiency analysis across four datasets. \textbf{Top:} Token consumption (K) at the best validation step. \textbf{Bottom:} Best step number (lower = faster convergence). aPSF consistently reaches peak validation with the fewest tokens and earliest steps.}
\label{tab:token_best_step}
\end{table}

aPSF reaches peak validation with 36--55\% fewer tokens than OPRO and ZERA across all datasets, confirming that the efficiency gains observed on MultiArith generalize broadly.

\section{Optimization Trajectory Analysis}
\label{app:trajectory}

\begin{figure}[h]
\centering
\includegraphics[width=0.95\linewidth]{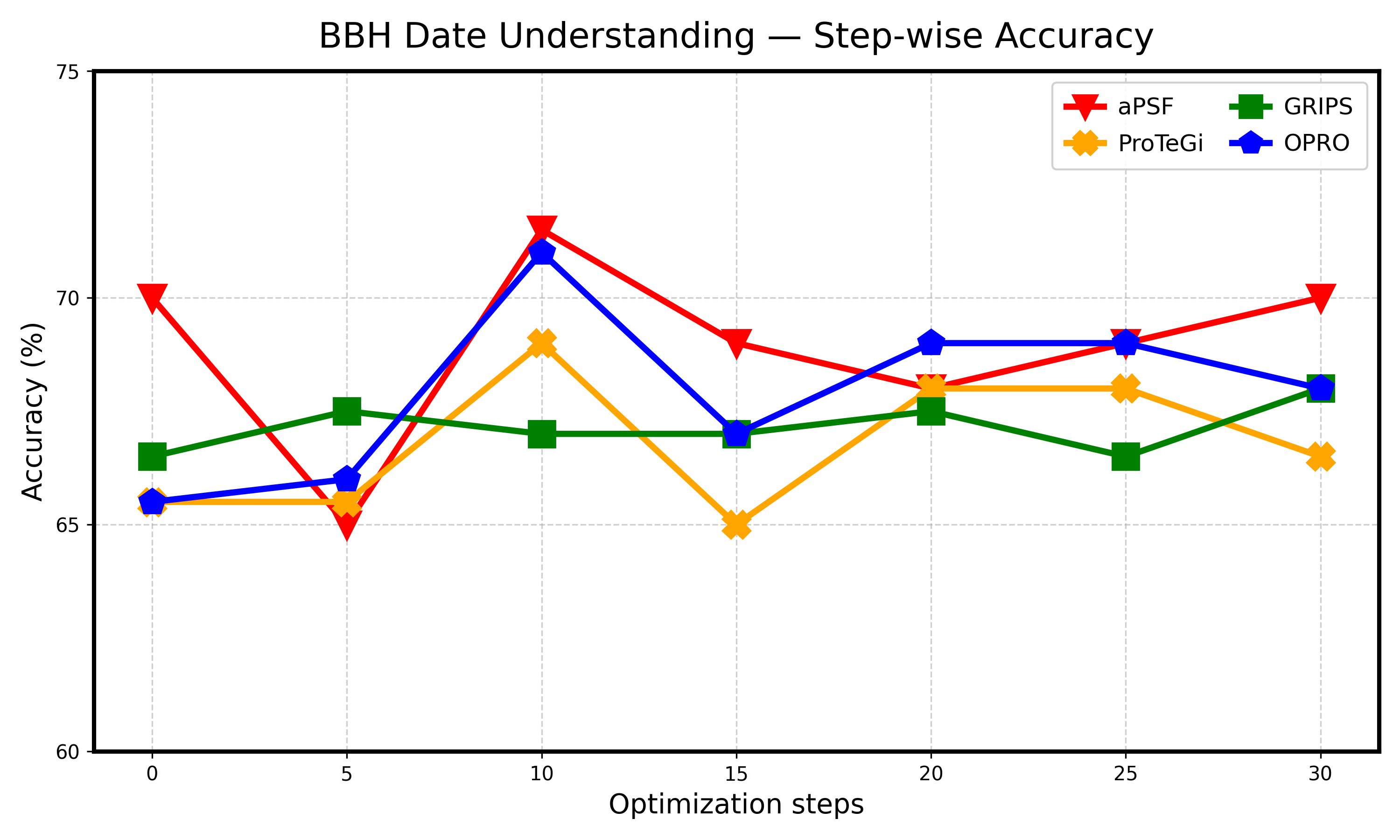}
\caption{Step-wise optimization trajectory on BBH Date Understanding. \textbf{aPSF (red)} achieves the highest peak accuracy ($>$71\% at Step 10) and maintains a leading position by Step 30. In contrast, monolithic methods like OPRO (blue) and ProTeGi (orange) exhibit higher volatility or lower convergence ceilings. \textit{Extended-budget analysis (30 steps) for visualization only; main results use $T_{\max}=10$. All methods use the same extended budget.}}
\label{fig:bbh_date_stepwise}
\end{figure}

We analyze the step-wise optimization process on the \textit{BBH Date Understanding} task. Figure~\ref{fig:bbh_date_stepwise} illustrates the validation accuracy trajectory over 30 steps.

\textbf{Key Observation:} aPSF demonstrates rapid convergence, reaching its peak performance early (Step 10). The acceptance threshold $\delta$ ensures monotonic progress on validation accuracy---rejected candidates cause plateau periods (e.g., Steps 5--8) rather than regressions. The error-guided mechanism then identifies productive factors to resume improvement, ultimately stabilizing at a higher accuracy level than baselines like GrIPS (Green) or ProTeGi.

\begin{figure*}[h]
\centering
\fbox{
    \begin{minipage}{0.96\textwidth}
        \small
        \textbf{Step 1: Error Diagnosis (Snapshot)} \\
        \textit{The Architect analyzes validation failures and maps them to specific factors.}
        \vspace{4pt}
        
        \textbf{[Sample 7]} \textbf{Error:} Misinterpreted algebraic expression. $\to$ \textcolor{blue}{Factor1\_UnderstandProblem} (Conf: 0.92) \\
        \textbf{[Sample 11]} \textbf{Error:} Incorrect probability independence assumption. $\to$ \textcolor{red}{Factor2\_SolveProblem} (Conf: 0.92) \\
        \textbf{[Sample 31]} \textbf{Error:} Output lengthy explanation instead of single letter. $\to$ \textcolor{orange}{Factor4\_OutputAnswer} (Conf: 0.93) \\
        \textbf{[Sample 43]} \textbf{Error:} Used incorrect LCM method for rate problem. $\to$ \textcolor{red}{Factor2\_SolveProblem} (Conf: 0.93) \\
        \textit{... (analyzed 15 error samples in total)}
        
        \vspace{8pt}
        \hrule
        \vspace{8pt}
        
        \textbf{Step 2: Aggregation \& Selection} \\
        \textit{The system aggregates error counts to identify the critical bottleneck.}
        
        \begin{itemize}[itemsep=0pt]
            \item \textbf{Factor1\_UnderstandProblem:} 3 errors (20.0\%)
            \item \textbf{Factor2\_SolveProblem:} \textbf{6 errors (40.0\%)} $\leftarrow$ \textbf{SELECTED BOTTLENECK}
            \item \textbf{Factor3\_SelectChoice:} 3 errors (20.0\%)
            \item \textbf{Factor4\_OutputAnswer:} 3 errors (20.0\%)
        \end{itemize}
        \textbf{Decision:} Prioritize optimizing \textbf{Factor2\_SolveProblem} due to highest error frequency.
        
        \vspace{8pt}
        \hrule
        \vspace{8pt}
        
        \textbf{Step 3: Candidate Generation} \\
        \textit{The Architect generates targeted refinements for Factor 2.}
        
        \textbf{Old Factor 2:} "Solve the problem step by step." \\
        \textbf{New Candidates:}
        1. "Work through the problem methodically, applying relevant concepts and checking equations..."
        2. "Solve the problem by clearly stating the needed formulas, performing calculation..."
        3. "Break the question into parts, apply appropriate reasoning, compute intermediate..."
        4. "Use step-by-step deduction: identify givens, apply relevant rules, simplify, and..."
    \end{minipage}
}
\caption{Real-world execution trace of the Error-Guided Factor Selection mechanism. The system identifies that 40\% of errors stem from reasoning flaws (\textsc{SolveProblem}), correctly prioritizing it over format or understanding issues for this iteration.}
\label{fig:execution_trace}
\end{figure*}

\section{Qualitative Analysis: Success and Failure Cases}
\label{app:qualitative}

\begin{figure*}[t] 
\centering
\fbox{
    \begin{minipage}{0.96\textwidth}
        \small
        \vspace{5pt}
        
        \textbf{Case 1: Integer Constraints on Quadratics (Sample 48)} \hfill \textcolor{darkgreen}{\textbf{[Success]}} \\
        \textit{Demonstrates rigorous algebraic manipulation and case enumeration.}
        \vspace{2pt}
        
        \textbf{Question:} For how many integer values of $a$ does the equation $x^2 + ax + 5a = 0$ have integer solutions for $x$? \\
        \textbf{aPSF Derivation (Summary):} 
        1. Applied Vieta's formulas: $p+q = -a, pq = 5a$.
        2. Derived the factorization equation: $pq + 5p + 5q + 25 = 25 \implies (p+5)(q+5) = 25$.
        3. Enumerated all factor pairs of 25: $(1,25), (-1,-25), (5,5), (-5,-5), (25,1), (-25,-1)$.
        4. Solved for $a$ in each case and filtered distinct values: $\{-16, 36, 0, 20\}$. \\
        \textbf{aPSF Answer:} \boxed{4} \quad \textbf{Ground Truth:} 4 \textcolor{green}{\cmark} \\
        \textbf{Analysis:} The structured prompt successfully guided the model to perform a variable transformation and exhaustive enumeration without hallucinating invalid pairs.
        
        \vspace{5pt}
        \hrule
        \vspace{5pt}
        
        \textbf{Case 2: Complex Roots of Unity (Sample 49)} \hfill \textcolor{darkgreen}{\textbf{[Success]}} \\
        \textit{Demonstrates handling of complex number properties and polynomial coefficients.}
        \vspace{2pt}
        
        \textbf{Question:} Let $\omega^7 = 1, \omega \ne 1$. $\alpha = \omega + \omega^2 + \omega^4$, $\beta = \omega^3 + \omega^5 + \omega^6$. Find $(a,b)$ for $x^2 + ax + b = 0$. \\
        \textbf{aPSF Derivation (Summary):}
        1. Identified property $\sum_{k=0}^6 \omega^k = 0 \implies \sum_{k=1}^6 \omega^k = -1$.
        2. Calculated sum: $\alpha + \beta = \sum_{k=1}^6 \omega^k = -1 \implies a = -(\text{sum}) = 1$.
        3. Calculated product: $\alpha\beta = (\omega + \omega^2 + \omega^4)(\omega^3 + \omega^5 + \omega^6)$. Expanded terms and simplified using $\omega^7=1$ to find $\alpha\beta = 2 \implies b = 2$. \\
        \textbf{aPSF Answer:} \boxed{(1, 2)} \quad \textbf{Ground Truth:} (1, 2) \textcolor{green}{\cmark} \\
        \textbf{Analysis:} The decomposition allowed the model to maintain clarity during the complex expansion of $\alpha\beta$, correctly identifying it as a sum of specific roots plus an integer.
        
        \vspace{5pt}
        \hrule
        \vspace{5pt}
        
        \textbf{Case 3: Nested Complex Functions (Sample 47)} \hfill \textcolor{red}{\textbf{[Failure]}} \\
        \textit{Demonstrates limitation in handling high-complexity functional composition.}
        \vspace{2pt}
        
        \textbf{Question:} Let $f(z)= \frac{z+a}{z+b}, |a|=1$. If $g(z)=f(f(z))$ and $g(g(z))=z$, find the difference between max and min values of $|b|$. \\
        \textbf{aPSF Derivation (Summary):}
        1. Correctly attempted to expand $g(z) = f(f(z))$ and $g(g(z))$.
        2. The algebraic expansion of the fourth-order composition $g(g(z))$ became overly complex.
        3. Incorrectly simplified the condition to a rotation argument ($f(z)$ must be a rotation by $\pm \pi/2$), concluding $|b|=1$ and difference is 0. \\
        \textbf{aPSF Answer:} \boxed{0} \quad \textbf{Ground Truth:} $\sqrt{3}-1$ \textcolor{red}{\xmark} \\
        \textbf{Analysis:} While the prompt structure enforced step-by-step derivation, the sheer algebraic complexity of expanding a nested Möbius transformation overwhelmed the model's context tracking, leading to a hallucinatory simplification. This suggests a need for a specialized \textsc{ToolUse} factor (e.g., SymPy) for such tasks.
        
        \vspace{5pt}
    \end{minipage}
}
\caption{Qualitative comparison of reasoning traces. \textbf{Cases 1 \& 2} show aPSF's ability to handle structured algebraic and number-theoretic reasoning. \textbf{Case 3} highlights a failure mode where extreme algebraic complexity leads to incorrect simplification, pointing to future directions for tool-augmented factors.}
\label{fig:qualitative_cases}
\end{figure*}

We analyze three representative samples (two successes, one failure) from the evaluation logs to illustrate the reasoning capabilities and limitations of aPSF. \textbf{Figure~\ref{fig:qualitative_cases}} presents the detailed breakdown.

\section{Limitations and Future Work}
\label{app:limitations}

\subsection{Current Limitations}

\paragraph{Architect quality dependency.}
aPSF's factorization quality depends on the Architect LLM's capability. Weaker architects may:
\begin{itemize}[leftmargin=*, itemsep=2pt]
\item Propose redundant factors (e.g., separate ``Instruction'' and ``Task Description'')
\item Under-discover useful factors (missing Format/Verifier on structured tasks)
\item Create overlapping factor boundaries (non-orthogonal decomposition)
\end{itemize}
Changing the Architect model can affect performance (Table~\ref{tab:math_main_grouped}).

\paragraph{Interventional scoring underestimates synergy.}
By isolating single-factor updates, interventional factor-level scoring may miss joint improvements
where two factors help only in combination (e.g., synergy between Examples and Rationale).
Future work could explore block-wise joint edits for factor pairs with high interaction.

\paragraph{LLM-based error diagnosis.}
\errorguided{} selection relies on an LLM to analyze step-level errors and map them to factors. While empirically effective (Table~\ref{tab:scheduler_gsm8k_aqua_ranked}), diagnostic quality depends on the LLM's reasoning capability and may introduce noise or bias. Future work could explore hybrid approaches combining error-pattern matching with learned factor-error associations.

\paragraph{Hyperparameter sensitivity.}
Performance can vary with the acceptance threshold $\delta$ and candidate count $N$; we use $\delta{=}1/|\mathcal{D}_{\mathrm{val}}|$ and $N{=}4$ by default (see Section~\ref{sec:setup}). Sensitivity analysis is provided in Appendix~\ref{app:hyperparam}.

\paragraph{Text-only reasoning benchmarks.}
Our evaluation focuses on math and logical reasoning tasks with textual I/O. Generalization to:
\begin{itemize}[leftmargin=*, itemsep=2pt]
\item \textbf{Multimodal tasks}: Vision-language tasks may require different factor types (e.g., \textsc{ImageGuidance})
\item \textbf{Tool-augmented settings}: Code execution or API calls may need \textsc{ToolSelection} and \textsc{ErrorHandling} factors
\item \textbf{Long-context tasks}: Summarization/QA over long documents may benefit from \textsc{ChunkingStrategy} factors
\end{itemize}

\subsection{Future Directions}

\paragraph{Adaptive candidate count.}
Currently $N{=}4$ candidates per update for all factors. Future work could adjust $N$ based on factor type (e.g., $N{=}8$ for Examples, $N{=}2$ for Format) or factor saturation (reduce $N$ after error analysis indicates diminishing returns).

\paragraph{Hierarchical factorization.}
Complex tasks may benefit from hierarchical factor structures (e.g., \textsc{Rationale} $\to$ \{\textsc{PlanningStep}, \textsc{ExecutionStep}, \textsc{VerificationStep}\}). Extending aPSF to multi-level factorization is an interesting direction.

\paragraph{Human-in-the-loop refinement.}
Allow practitioners to inspect discovered factors, manually adjust boundaries, or freeze critical factors (e.g., compliance-sensitive Verifier). 

\paragraph{Improved error diagnosis models.}
Develop specialized models for mapping error patterns to factors, potentially fine-tuned on factor-error association data. This could improve selection accuracy beyond generic LLM-based diagnosis.

\end{document}